\documentclass[11pt]{article}
\usepackage{Basic}
\usepackage{todonotes}

\newcommand{\Lip}{{\rm Lip}}

\allowdisplaybreaks

\title{Stein's Method for Marginals on Large Graphical Models}
\author{Tiangang Cui\thanks{School of Mathematics and Statistics, The University of Sydney (\texttt{tiangang.cui@sydney.edu.au})} \and Shuigen Liu\thanks{Department of Mathematics, National University of Singapore (\texttt{shuigen@u.nus.edu}, \texttt{xin.t.tong@nus.edu.sg})} \and Xin T.~Tong\footnotemark[2]}
\date{\today}

\begin{document}
\maketitle

\begin{abstract}
Many spatial models exhibit locality structures that effectively reduce their intrinsic dimensionality, enabling efficient approximation and sampling of high-dimensional distributions. 
However, existing approximation techniques primarily focus on joint distributions and do not provide precise accuracy control for low-dimensional marginals, which are of primary interest in many practical scenarios.
By leveraging locality structure, we establish a dimension-independent uniform error bound for the marginals of approximate distributions.
Inspired by Stein's method, we introduce a novel $\delta$-locality condition that quantifies the locality in distributions, and link it to structural assumptions such as sparse graphical models. 
The theoretical guarantee motivates the localization of existing sampling methods, as we illustrate through the localized likelihood-informed subspace method and localized score matching. 
We show that by leveraging the locality structure, these methods greatly reduce the sample complexity and computational cost via localized and parallel implementations.
\end{abstract}

\bigskip
\noindent\textbf{Keywords:} Functional inequality, Stein's method, high-dimensional distributions, graphical models

\bigskip
\noindent\textbf{Mathematics Subject Classification:}  60E15, 60B10, 62F15, 62H22

\section{Introduction}
\label{sec:Intro}

Approximating and sampling from probability distributions are fundamental tasks in computational mathematics \cite{MR3274599,24M1694781,chen2026convergence}, statistics \cite{MR964183,chewi2023logconcave,MR373208}, generative modeling \cite{kingma2013auto,NIPS2014_5ca3e9b1,NEURIPS2020_4c5bcfec}, statistical physics \cite{MR373208,clifford1990markov}, and beyond.
A central challenge in these tasks arises from high dimensionality: the complexity of most approximation and sampling algorithms scales poorly with the dimension of random variables.
The design of efficient algorithms often relies on exploiting the intrinsic structures of the target distribution to mitigate this dimensionality challenge.
Examples include low-dimensional subspaces \cite{MR3274599,MR4474562}, sparsity \cite{10.5555/2834535}, and the manifold assumption \cite{kingma2013auto,NIPS2014_5ca3e9b1}.

In this paper, we study a general class of intrinsic structures, referred to as \emph{locality}, which naturally arises in many real-world phenomena, for instance, interactions among a subset of genes from the genome forming a biological pathway, short-range interactions between non-bonded atoms in molecular liquids or gases, and the spatial and/or temporal decay of energy propagation in measurement processes due to diffusion and scattering.
Systems with such locality behavior often exhibit sparse dependencies among their states \cite{MR373208,MR3375889}, a key feature that underpins the modeling and efficient simulation of complex systems across various fields, including spatial statistics \cite{MR373208,MR964183,MR3538706}, data assimilation \cite{2001MWRv..129..123H,2001MWRv129.7690,MR3375889}, quantum mechanics \cite{PhysRevLett.76.3168,MR3032838}, and numerical solution to partial differential equations \cite{MR2054351}.
Our goal is to introduce a novel quantification of locality in probability distributions and to develop corresponding scalable approximation and sampling methods that are robust to dimensionality.

\subsection{Functional inequalities via Stein's method}
\label{sec:stein_intro}

Let $\mcP_1(\mR^d)$ denote the collection of differentiable probability density functions on $\mR^d$, with respect to the Lebesgue measure, corresponding to random variables with finite first moments. For $\bv\in\mR^d$, let $\norm{\bv}_p$ denote the vector $p$-norm and let $\norm{\bv} = \norm{\bv}_2$ denote the Euclidean norm. For a matrix $A$, let $\norm{A}$ denote the matrix norm induced by the vector $2$-norm.
We consider a target density $\pi \in \mcP_1(\mR^d)$ approximated by $\pi'\in \mcP_1(\mR^d)$ using a numerical approximation scheme or a sampling method, and quantify the approximation error using the $1$-Wasserstein distance,
\begin{equation}
    \sfW_1(\pi,\pi') = \inf_{\gamma\in \Pi(\pi,\pi')} \int_{\mR^d} \norm{\bx-\by} \mdd \gamma(\bx,\by),
\end{equation}
where $\Pi(\pi,\pi')$ denotes the set of couplings of $\pi$ and $\pi'$, i.e., the marginals of each joint density $\gamma \in \Pi(\pi,\pi')$ are $\pi$ and $\pi'$, respectively, such that $(\bX,\bY) \sim \gamma$ implies $\bX \sim \pi, \bY\sim \pi'$.
Our framework is centered on the gradient control of $\sfW_1(\pi,\pi')$ via \emph{Stein's method} \cite{MR402873,MR882007}.
In this work, we will use the following particular Stein operator, referred to as the \emph{Langevin Stein operator} in \cite{MR4019878}.

\begin{defn}[Stein operator]
\label{def:stein_opt}
Let $\pi \in \mcP(\mR^d)$ be a probability distribution with positive $C^1$ density.
We define its associated \emph{Stein operator} $\sfL_\pi: C^2_{\rm c}(\mR^d) \to C_{\rm c}(\mR^d)$ as
\begin{equation}
\label{eqn:stein_opt}
    \sfL_\pi u (\bx) := \Delta u(\bx) + \nabla \log \pi(\bx) \cdot \nabla u(\bx). 
\end{equation}
The domain of $\sfL_\pi$ can be extended  via duality to the weighted Sobolev space
\begin{equation}
\label{eqn:H1}
    H^1(\pi) = \left\{ u \in W^{1,2}_{\rm loc}(\mR^d) : \int \norm{\nabla u(\bx)}^2 \pi(\bx) \mdd \bx < \infty \right\}.
\end{equation}
In this setting, we can define the extended operator, $\sfL_\pi : H^1(\pi) \to H^1(\pi)^\ast$, in the weak form:
\begin{equation}
\label{eqn:stein_opt_dual}
    \int \sfL_\pi u(\bx) v(\bx) \pi(\bx) \mdd \bx = - \int \nabla u(\bx) \cdot \nabla v(\bx) \pi(\bx) \mdd \bx, \quad \forall v \in H^1(\pi).
\end{equation}
\end{defn}

Most of our results are built on the following transport-type inequality.

\begin{prop}
\label{prop:OVineqn_full}
Suppose a target density $\pi$ is $m$-strongly log-concave, i.e., $- \nabla^2 \log \pi(\bx) \succeq m I_d$ for all $\bx \in \mR^d$ for some $m>0$.
Here $A\succeq B$ denotes the Loewner order, i.e., $A-B$ is positive semi-definite.
The $1$-Wasserstein distance between the target $\pi$ and the approximation $\pi'$ satisfies the upper bound
\begin{equation}
\label{eqn:OVineqn_full}
    \sfW_1(\pi,\pi') \le C_\pi \E_{\pi'} \Rectbrac{ \|\nabla \log \pi-\nabla \log \pi' \| },
\end{equation}
for a constant $C_\pi$ depending on $m$.
\end{prop}

\begin{proof}[Proof sketch]
Let $\Lip_1(\mR^d)$ denote the class of $1$-Lipschitz functions. By Kantorovich duality \cite{MR2459454},\begin{equation}
    \sfW_1(\pi,\pi') = \sup_{\phi \in \Lip_1(\mR^d)} \int \phi (\bx) \Brac{ \pi(\bx) - \pi' (\bx)} \mdd \bx = \sup_{\phi \in \Lip_1(\mR^d)} \int \big(  \E_{\pi}[\phi(\bX)] - \phi (\bx) \big) \pi'(\bx) \mdd \bx.
\end{equation}
For any $\phi \in \Lip_1(\mR^d)$, given $u_\phi(\bx)$ the solution to the \emph{Stein equation}
\begin{equation}
    \sfL_{\pi} u_\phi (\bx) = \phi(\bx) - \E_{\pi} [\phi(\bX)],
\end{equation}
and applying integration by parts \eqref{eqn:stein_opt_dual}, we have 
\begin{equation}
\begin{split}
    \int \big(  \E_{\pi}[\phi(\bX)] - \phi (\bx)\big) \pi'(\bx) \mdd \bx & = \int \nabla u_\phi(\bx) \cdot \Brac{ \nabla \log \pi' (\bx) - \nabla \log \pi (\bx) } \pi'(\bx) \mdd \bx \\
    & \le \sup_{\bx} \big\| \nabla u_\phi(\bx) \big\|  \; \E_{\pi'} \Rectbrac{ \big \| \nabla \log \pi' (\bx) - \nabla \log \pi (\bx) \big\| }.
\end{split}
\end{equation}
Defining the constant $C_\pi := \sup_{\phi\in \Lip_1(\mR^d)} \sup_{\bx} \big\| \nabla u_\phi(\bx) \big\|$ for all solutions to the Stein equation, the inequality follows.
The constant $C_\pi$ is called the \emph{Stein factor} \cite{MR3548768,MR4019878,MR4583066}, and satisfies $C_\pi \le m^{-1}$ for $m$-strongly log-concave densities.
\end{proof}

The work of \cite{MR4583066} gives an explicit derivation of \eqref{eqn:OVineqn_full} via Stein's method. In fact, the inequality follows directly from bounding the Stein factor, which is already established in \cite{MR3548768,MR4019878}. 
We also refer readers to the review \cite{10.1214/22-STS863} for more details on applications of Stein's method.
The inequality \eqref{eqn:OVineqn_full} is closely related to Talagrand's inequality \cite{MR1392331}. A far-reaching extension was later obtained by Otto and Villani \cite{MR1760620}, who proved that if $\pi$ is $m$-strongly log-concave, then for any $\pi' \in \mcP_2(\mR^d)$,
\begin{equation}
    \sfW_2^2(\pi,\pi') \le \frac{2}{m} \KL(\pi'\|\pi) \le \frac{1}{m^2} \E_{\pi'} \Rectbrac{ \|\nabla \log \pi-\nabla \log \pi' \|^2 },
\end{equation}
where $\KL(\pi'\|\pi) = \E_{\pi'} [ \log \frac{\pi'}{\pi} ] $ is the Kullback--Leibler divergence.
Compared with the Otto--Villani inequality, the inequality \eqref{eqn:OVineqn_full} replaces the $2$-Wasserstein distance with the $1$-Wasserstein distance, and correspondingly replaces the $L^2$-norm on the right-hand side with the $L^1$-norm.
Such an inequality is also referred to as a transport--information inequality in \cite{Guillin2009,10.1214/19-AAP1530}, where a set of sufficient and necessary conditions are discussed.
For further discussion and background on transport-type inequalities, we refer to the survey of \cite{MR2895086}.

Note that the right-hand side of the inequality \eqref{eqn:OVineqn_full} involves the difference between the gradients of the log-densities (i.e., the \emph{score functions}), which serves as a loss function in many density estimation and approximation methods, e.g., score matching \cite{JMLR:v6:hyvarinen05a,NEURIPS2019_3001ef25} and diffusion models \cite{NEURIPS2020_4c5bcfec}.
Since the score functions are $d$-dimensional vectors, the right-hand side of \eqref{eqn:OVineqn_full} typically grows like $\sqrt{d}$. This suggests that higher-dimensional problems may require more training data or increased computational effort to compensate for such growth.
We aim to overcome this pessimistic scaling behavior by exploiting the locality structure of the target distribution.

\subsection{Roadmap: controlling marginal errors using \texorpdfstring{$\delta$}{delta}-locality}
\label{sec:roadmap}

To facilitate the quantification of locality, we decompose the random variables modeling the states of these systems into local blocks.
Given a $d$-dimensional space, we partition the full coordinate set $[d] := \{1,\dots,d\}$ into $K$ disjoint subsets, $\mcI_1, \ldots, \mcI_K \subseteq [d]$ such that $[d] = \cup_{j = 1}^K \mcI_j$, and $\forall j \neq k, ~ \mcI_j \cap \mcI_k = \emptyset$.
Let $d_j := |\mcI_j|$ denote the cardinality of $\mcI_j$. For each coordinate subset, $\mcI_j = \{j_1,\dots,j_{d_j}\}$, we denote the corresponding subset of random variables by 
\begin{equation}
    \bX_{\mcI_j} = ( X_{j_1}, X_{j_2}, \dots, X_{j_{d_j}} ) \in \mR^{d_j}, 
\end{equation}
and use the shorthand notation $\bX_{j} := \bX_{\mcI_j}$ when there is no ambiguity. This way, the full set of random variables $\bX$ can be decomposed as $\bX = (\bX_1, \ldots, \bX_K)$.
A special case is the singleton partition, where each subset contains exactly one component of $\bX$, that is, $\mcI_j = \{j\}$ and $\bX_j = X_j$, the $j$-th element of $\bX$. Given the complement coordinate subset $\mcI_j^c = [d] \setminus \mcI_j^{}$, we introduce the shorthand notation $\bX_{-j} := \bX_{\mcI^c_j}$ such that $\bX = (\bX_j, \bX_{-j})$. For $\bX \sim \pi$, we write the marginal density of $\bX_j$ as
\begin{equation}
    \pi_j (\bx_j) = \int \pi(\bx_j,\bx_{-j}) \mdd \bx_{-j}.
\end{equation}
The gradient of a function $f : \bx \rightarrow f(\bx)$ with respect to a subset of variables $\bx_{j}$ is denoted by $\nabla_j f := \nabla_{\bx_j} f: \mR^{d} \rightarrow \mR^{d_j}$, and the second derivative of $f$ with respect to subsets of variables $\bx_{j}$ and $\bx_{k}$ are denoted by $\nabla_{jk}^2 f = \nabla_j^{} \nabla_k\matT f := \nabla_{\bx_j} \Brac{ \nabla_{\bx_k} f }\matT : \mR^{d} \rightarrow \mR^{d_j \times d_k}$.

Besides the weighted Sobolev space $H^1(\pi)$ \eqref{eqn:H1}, we also define the centered subspace
\begin{equation}
    L_0^2(\pi) = \left\{ u \in L^2(\pi) : \int_{\mR^d} u(\bx) \pi(\bx) \mdd \bx = 0 \right\}, \quad H_0^1(\pi) = H^1(\pi) \cap L_0^2(\pi).
\end{equation}
Throughout the paper, we assume that $\pi$ satisfies a Poincar\'e inequality: there exists $C_P < \infty$ such that for any $u \in H_0^1(\pi)$, $\E [u^2] \le C_P \E [\|\nabla u\|^2]$.
It is easy to see that, by the Lax--Milgram theorem, the Stein operator $\sfL_\pi$ defined in \eqref{eqn:stein_opt_dual} is invertible in $H_0^1(\pi)$.

\begin{defn}[$\delta$-locality]
\label{def:PoisGrad}
Let $\pi \in \mcP(\mR^d)$ have a positive $C^1$ density and satisfy a Poincar\'e inequality.
We say that $\pi$ is {\bf $\delta$-localized} for some dimension-free constant $\delta>0$ if, for every $i\in[K]$ and every $1$-Lipschitz function $\phi : \mR^{d_i} \gto \mR$, the solution $u(\bx)$ to the \emph{marginal Stein equation}
\begin{equation}
\label{eqn:PoisEqn}
    \sfL_\pi u (\bx) := \Delta u(\bx) + \nabla \log \pi(\bx) \cdot \nabla u(\bx) = \phi(\bx_i) - \E_\pi [\phi(\bx_i)]
\end{equation}
exists in $H_0^1(\pi)$ and satisfies the gradient bound
\begin{equation}
\label{eqn:PoisGradBound}
    \norm{ \nabla u }_{\infty,1} := \sum_{j=1}^K \norm{ \nabla_j u }_{L^\infty} \le \delta.
\end{equation}
Here $\norm{ \nabla_j u }_{L^\infty} := \sup_{\bx\in\mR^d} \norm{ \nabla_j u(\bx) }$ denotes the $L^\infty$ norm of $\norm{\nabla_j u}_2$.
\end{defn}

Note the solution to the marginal Stein equation \eqref{eqn:PoisEqn} exists uniquely in $H_0^1(\pi)$ since the right-hand side of \eqref{eqn:PoisEqn} lies in $H_0^1(\pi)$.
We aim to analyze how accurate a target probability density  $\pi \in  \mcP_1(\mR^d)$ can be approximated by a $\delta$-localized density $\pi' \in  \mcP_1(\mR^d)$. Such localized approximations $\pi'$ form the basis for designing efficient numerical schemes, and the extent to which $\pi$ conforms to a $\delta$-localized structure directly determines the accuracy of those schemes. Since $\delta$-localized distributions arise naturally in many applications, we begin in \Cref{sec:LocalDist} by establishing the $\delta$-locality condition for a range of settings, including sparse graphical models, and diagonally dominant distributions.

An important consequence of the $\delta$-locality condition is that perturbations of these distributions induce only localized, dimension-independent errors on the marginal distributions.
In \Cref{sec:MarginTala}, we show in \Cref{thm:margin} that such errors on the marginal distributions can be controlled by a gradient bound 
\begin{equation}
\label{eqn:Margin}
    \max_{i\in [K]} \sfW_1(\pi_i^{},\pi_i') \le \delta \cdot \max_{j\in[K]} \int \norm{ \nabla_j \log \pi (\bx) - \nabla_j \log \pi' (\bx) } \pi(\bx) \mdd \bx,
\end{equation}
via the marginal Stein equation \eqref{eqn:PoisEqn}.
The bound \eqref{eqn:Margin} yields two practical benefits.
First, the approximation error of low-dimensional marginals can be made independent of the ambient dimension, thereby enabling efficient approximation and sampling methods; see \cite{MR3901708,MR4111677,24M1694781,chen2026convergence,lacker2025hierarchical} for examples.
Second, the bound is computable in practice, since it requires only the evaluation of the gradient of the score function, rather than the often inaccessible marginal distributions.
In \Cref{sec:MarginTala}, we also show a natural extension of the bound \eqref{eqn:Margin} to marginal distributions of multiple blocks of coordinates.

\begin{rem}
The inequality \eqref{eqn:Margin} can be viewed as a marginal analogue of the bound on the $1$-Wasserstein distance in \eqref{eqn:OVineqn_full}: the left-hand side quantifies discrepancies between the marginals of the distributions of interest, while the right-hand side captures differences in the individual components of the score.
Compared to \eqref{eqn:OVineqn_full}, the bound in \eqref{eqn:Margin} yields a dimension-independent, uniform control of marginal errors.
Standard transport inequalities can already yield dimension-independent bounds on the average of marginal discrepancies, such as \(\frac{1}{K}\sum_{i=1}^K \sfW_1(\pi_i,\pi_i')\); see \cite{lacker2025hierarchical}. 
Our result instead controls the worst marginal discrepancy, $\max_i \sfW_1(\pi_i,\pi_i')$. Such a uniform guarantee is strictly stronger than average-error control and cannot generally be recovered from it.
This is particularly valuable in settings where each marginal corresponds to a distinct location in spatial data \cite{MR4580673} or manifold data \cite{10.3150/24-BEJ1804,khoo2024temporal}.
\end{rem}

%{TC will do this. \color{red} \Cref{sec:LLIS} LIS

%Moreover, the marginal form in \eqref{eqn:Margin} suggests that to obtain a uniform bound on the marginal error, it suffices to minimize the worst coordinate error of the score function, which inspires the localized score matching in \Cref{sec:LSM}. In particular, we show in Theorem \ref{thm:LSM} the necessary training data can be dimension independent.}

Moreover, controlling the marginal errors in \eqref{eqn:Margin} enables decomposing high-dimensional computational tasks into local ones. In \Cref{sec:LLIS}, we build on this idea to develop a localized likelihood-informed subspace framework, which combines both low-rank structure and locality, for approximating high-dimensional probability densities that commonly arise in inference problems. In \Cref{sec:LSM}, we consider another practical setting, namely score-based density estimation, in which the target probability density is not directly accessible and only samples from the target distribution are available. In this scenario, the scalability of sample complexity with respect to dimension becomes the bottleneck. Our results show that, under locality assumptions, one can obtain essentially dimension-independent sample complexity.

\subsection{Related applications of locality}
\label{sec:related_work}

In spatial statistics, the Vecchia approximation \cite{MR964183} approximates Gaussian processes by discarding long-range conditional dependencies; that is, a conditional density of the form \(\pi(\bx_{j} \mid \bx_{-j})\) in the Gaussian process is replaced by \(\pi(\bx_{j} \mid \bx_{\mcN(j)})\), where $\mcN(j)$ denotes a small set of neighboring variable blocks. This significantly reduces the computational complexity. Various extensions building on this nearest-neighbor approximation idea have since been developed; see \cite{MR3538706} for a concise review.

In data assimilation, a naïve implementation of the ensemble Kalman filter requires a large ensemble size to accurately estimate sample covariances in high-dimensional problems.
Localization methods \cite{2001MWRv..129..123H,2001MWRv129.7690} have therefore become essential tools for mitigating this large-sample-size bottleneck.
For example, covariance localization suppresses spurious long-range correlations in the sample covariance $\widehat{\sfC}$ through a Hadamard product,
\(
    \widehat{\sfC}_{\mathrm{loc}} = \rho \circ \widehat{\sfC}
\)
where $ \rho_{ij} = \psi(|i-j|) $ is the $(i,j)$-th element of the localization matrix defined by a rapidly decaying taper function $\psi : \mN \to \mR_{+}$.
Such localization techniques have been shown to substantially reduce sampling errors and improve filter performance \cite{2001MWRv..129..123H,2001MWRv129.7690,MR3375889}.

In quantum mechanics, the nearsightedness principle \cite{PhysRevLett.76.3168} states that the properties of a quantum system are primarily determined by local interactions and are insensitive to distant perturbations.
This principle can be formalized through the exponential decay of the density operator $\rho(\br, \br')$ with respect to the physical distance $\|\br - \br'\|$ \cite{PhysRevLett.76.3168,MR3032838}, which enables efficient sparse approximations of the density operator.
Such approximations can be viewed as a quantum analogue of the localization of probability distributions, and they dramatically reduce the computational cost of tasks such as computing ground states or evaluating expectation values of observables.
The principle also underpins the theoretical foundations of interatomic potentials used in molecular dynamics simulations \cite{MR3463688}.

In recent years, there has been rapidly growing interest in sampling methodologies that exploit locality structures \cite{MR3901708,MR4108690,MR4111677,MR5004938,24M1694781}.
\cite{MR3901708} introduces a localization technique for Markov chain Monte Carlo in Bayesian inverse problems and proposes a localized Metropolis-within-Gibbs sampler.
This idea is extended in \cite{MR4111677} to a Metropolis-adjusted Langevin-within-Gibbs sampler with a dimension-independent convergence rate under localized distributions, and further extended in \cite{MR4108690} to parallelized proposal computation.
More recently, \cite{24M1694781} leverages localized structure to obtain dimension-independent approximation errors for marginals in an image deblurring problem; their analysis forms a foundation on which we generalize the idea in this paper.
In the context of unadjusted Langevin algorithms, \cite{chen2026convergence,lacker2025hierarchical} shows that when the target distribution exhibits locality, the marginal bias of the unadjusted Langevin algorithm grows at most logarithmically with dimension.

Beyond Markov chain Monte Carlo, a message passing Stein variational gradient descent method is proposed in \cite{pmlr-v80-zhuo18a,pmlr-v80-wang18l}, which exploits conditional independencies to compute descent directions in a coordinate-wise manner, mitigating the degeneracy of kernel methods in high dimensions.
In a related development, \cite{MR5004938} introduces a localized version of the Schrödinger bridge sampler of \cite{rsta.2024.0332}, replacing one high-dimensional Schrödinger bridge problem with a number of lower-dimensional ones, thereby avoiding the exponential dependence of sample complexity on dimension.

\section{Examples of \texorpdfstring{$\delta$}{delta}-localized distributions}
\label{sec:LocalDist}

In this section, we derive the solution to the marginal Stein equation, establish the $\delta$-locality condition (Definition~\ref{def:PoisGrad}) for two classes of probability distributions, and discuss the connection between $\delta$-locality and gradient estimate of Langevin semigroups.

\subsection{Solution to the marginal Stein equation}
\label{sec:SoluMarginal}

Most of the results of this section rely on the following lemma, which provides an explicit form of the solution to the marginal Stein equation \eqref{eqn:PoisEqn} and its gradient estimate using the overdamped Langevin dynamics
\begin{equation}
\label{eqn:Langevin}
    \mdd \bX_t^\bx = \nabla \log \pi(\bX_t^\bx ) \mdd t + \sqrt{2} \mdd W_t , \quad \bX_0^\bx = \bx,
\end{equation}
where the generator of the Langevin dynamics, $\sfL_\pi = \Delta + \nabla \log \pi \cdot \nabla$, coincides with the Stein operator of \eqref{eqn:PoisEqn}.

\begin{lem}
\label{lem:ExpSolu}
Suppose $\pi$ is strongly log-concave. Consider $i$-th block of the random variable, $\bX_i$. For any $1$-Lipschitz function $\phi:\mR^{d_i} \to \mR$, the solution to the corresponding marginal Stein equation \eqref{eqn:PoisEqn} is given by
\begin{equation}
\label{eqn:Solu}
    u(\bx) =  \int u(\bx) \pi(\bx) \mdd \bx - \int_0^\infty \E \Rectbrac{ \phi(\bX_{t,i}^\bx) - \E_\pi[\phi(\bx_i)] } \mdd t,
\end{equation}
where $\bX_t^\bx $ is the path of the overdamped Langevin dynamics \eqref{eqn:Langevin}.
Note that the solution is unique up to an additive constant.
As a consequence, the following gradient estimate holds:
\begin{equation}
\label{eqn:GradCtrl}
    \norm{ \nabla_j u(\bx) } \le \int_0^\infty \E \Rectbrac{ \normo{ \nabla_{\bx_j} \bX_{t,i}^\bx } } \mdd t,
\end{equation}
where $\nabla_{\bx_j} \bX_{t,i}^\bx \in  \mR^{d_i\times d_j}$ denotes the partial derivative of the $i$-th block of the path of the Langevin dynamics at time $t$ with respect to the $j$-th block of the initial condition $\bx$.
\end{lem}

\noindent The proof is provided in \Cref{app:lem_ExpSolu}.

\subsection{Localized graphical models}
\label{sec:LGM}

Undirected graphical models \cite{MR2778120} are widely used for representing local interactions among random variables. We show here that undirected graphical models, also known as Markov random fields \cite{clifford1990markov}, are $\delta$-localized whenever their underlying graphs are effectively localized.

\subsubsection{Introduction to graphical models}
\label{sec:Intro_GM}

Consider a $d$-dimensional random vector $\bX$ partitioned into $K$ disjoint subsets,
\(
    \bX = (\bX_j)_{j \in [K]},
\)
and suppose each subset, $\bX_j$, is associated with a vertex $j$ of an undirected \emph{dependency graph} $\sfG = (\mathsf{V}, \mathsf{E})$, where $\mathsf{V} = [K]$. 
For simplicity, we assume that $\sfE$ contains all self-loops, i.e., $(j,j)\in \sfE$ for every $j \in \sfV$.

We denote the set of neighboring vertices of $j$ in the graph $\sfG$, including the vertex $j$ itself, by $\mcN(j) = \{k \in \sfV: (j,k)\in \sfE\} $, and let $\mcN'(j) = \mcN(j) \setminus \{j\}$ be the neighbor set with the self-vertex removed. We also define the order-$q$ neighbor set of a vertex $j\in \sfV$ by
\begin{equation}
    \mcN(j;q) := \{ k \in \sfV: \sfd_\sfG(j,k) \le q \}, \quad q \in \mN, 
\end{equation}
where $\sfd_\sfG(j,k)$ is the graph path distance between vertices $j,k\in \sfV$, i.e., the minimum number of edges along any path connecting them. For convenience, we set $\sfd_\sfG(j,j)=0$, and $\sfd_\sfG(j,k)=\infty$ when $j$ and $k$ lie in different connected components.
Note that $\mcN(j) \equiv \mcN(j;1)$ and $\mcN'(j) = \mcN(j; 1) \setminus \mcN(j;0)$ under this definition.

The edges of the graph $\sfG$ encode the local dependencies among elements of $\bX$. For every vertex $j \in \sfV$, each subset of variables \(\bX_i\) associated with a nonadjacent vertex $i \in \sfV \setminus \mcN(j)$ is conditionally independent of $\bX_j$ given the neighboring components of $\bX_j$, that is, 
\begin{equation}
\label{eqn:MRF}
    \bX_i \ci \bX_j ~|~ \bX_{\smash{\mcN'(j)}}, \quad \forall i \in \sfV \setminus \mcN(j),
\end{equation}
where $\ci$ denotes conditional independence, and $\bX_{\smash{\mcN'(j)}}$ contains elements of $\bX$ corresponding to the neighboring index set $\mcI_{\smash{\mcN'(j)}} = \cup_{k \in \smash{\mcN'(j)}} \,\mcI_k$.

For a twice differentiable density $\pi(\bx)$, conditional independence can be equivalently characterized through the second-order mixed derivatives \cite{MR4111677}:
\(
    \nabla_{ij}^2 \log \pi(\bx) = 0, \forall (i,j) \notin \sfE.
\)
Another equivalent characterization of the sparse graphical model is the \emph{clique factorization} \cite{clifford1990markov} of the log density, and in a slightly relaxed form, it reads
\(
    \log \pi(\bx) = \sum_{j\in [K]} u_j\big(\bx_{\mcN(j)}\big),
\)
for some functions $(u_j)_{j\in [K]}$.
The clique factorization is a key property that enables efficient approximation and computation of the distribution, as it implies that the log density can be decomposed into essentially low-dimensional functions. We will exploit this in \Cref{sec:LSM}.

\subsubsection{Graph locality}

We introduce the following definition to quantify the locality of the graph structure.

\begin{defn}
\label{def:loc_graph}
An undirected graph $\sfG$ is called $\boldsymbol{(S,\nu)}$\textbf{-local} if there exist $S,\nu \in \mZ_+$ such that the cardinality of the order-$q$ neighbor set satisfies
\begin{equation}
\label{eqn:loc_graph}
    |\mcN(j;q)| \le 1 + S q^\nu , \quad \forall j \in \sfV, ~ \forall q\in\mN,
\end{equation}
where $S$ controls the maximal size of immediate neighborhoods, and $\nu$ is the effective ambient dimension of the graph that controls the growth of the neighborhood volume as a function of the radius $q$.
\end{defn}

In the above definition, we require the maximal cardinality of the order-$q$ neighbor sets to grow at most polynomially in $q$ to ensure effective locality. A motivating example is the lattice on $\mZ^\nu$, which has a na\"ive bound on the neighborhood volume:
\(
    |\mcN(j;q)| = | \{k\in \mZ^\nu: \norm{k}_1 \le q\} | \le (2q+1)^\nu < 1 + (3q)^\nu.
\)
This way, the lattice on $\mZ^\nu$ is $(3^\nu,\nu)$-local.
We note that similar conditions appear in the analysis of delocalization of bias in Langevin algorithms \cite{chen2026convergence,lacker2025hierarchical}.

A common class of probability distributions equipped with such sparse dependency graphs is those with a (block) banded \emph{local precision matrix}, $\sfH(\bx) := -\nabla^2 \log \pi(\bx)$. Specifically, for some (block) bandwidth $B$, the local precision matrix satisfies
\(
    \nabla^2_{jk} \log \pi(\bx) = 0, \forall |j-k| > B.
\)
In this case, the corresponding dependency graph is $(2B,1)$-local. The following example shows the locality of a nearest-neighbor spin chain model.

\begin{exmp}[Nearest-neighbor spin chain]
A nearest-neighbor spin chain describes a chain of real-valued spins $\{x_j\}_{j=1}^{d}$, where each spin $x_j \in \mR$ interacts only through local nearest-neighbor couplings. Its Gibbs distribution is given by
\begin{equation}
    \pi(x) = \frac{1}{Z} \exp \Brac{ - \sum_{j=1}^{d} V(x_j) - \sum_{j=1}^{d-1} W(x_j,x_{j+1}) },
\end{equation}
where $V$ satisfies $0<m\le V''(x)\le M<\infty$, and $W(x,y) = \frac{\beta}{2} (x-y)^2$ with $\beta\ge0$ is the nearest-neighbor interaction.
Note the local precision matrix satisfies $mI_d\preceq-\nabla^2\log\pi\preceq(M+4\beta)I_d$. Consider the singleton partition, $\bX = (X_j)_{j \in [d]}$. The associated dependency graph is $\sfG = ([d], \sfE)$, where $\sfE = \{(j,j+1)\}_{j = 1}^{d-1} \cup \{(j,j)\}_{j = 1}^d$. The local precision matrix of the density has bandwidth $1$, and thus the dependency graph of the distribution is $(2,1)$-local.
\end{exmp}

\subsubsection{From graph locality to \texorpdfstring{$\delta$}{delta}-locality}
\label{sec:graph_to_loc}

\begin{thm}
\label{thm:MRF_loc}
Suppose $\pi \in \mcP(\mR^d)$ is associated with a $(S,\nu)$-local dependency graph $\sfG$. Suppose further there exist constants $m,M$ with $0<m\le M<\infty$ such that the local precision matrix of the density $\pi$ satisfies
\(
    m I_d \preceq - \nabla^2 \log \pi(\bx) \preceq M I_d, \forall \bx \in \mR^d.
\)
Then, the density $\pi$ is $\delta$-localized with $\delta = \frac{S \nu! \kappa^\nu }{m}$ where $\kappa = \frac{M}{m}$ is the condition number. 
\end{thm}

\begin{rem}
The condition number $\kappa$ plays a crucial role in establishing locality. It is well known that $\kappa$ determines whether banded structure is preserved under matrix inversion \cite{MR758197,MR2848418}; in probabilistic terms, for a Gaussian distribution with the covariance matrix $\sfC$, a moderate condition number ensures that local correlations (i.e., $\sfC$ being nearly banded) imply local conditional dependencies (i.e., $\sfC^{-1}$ being nearly banded).
Moreover, achieving a dimension-independent $\delta$ requires $\kappa$ to remain bounded as the dimension $d$ grows.
This assumption is realistic in many spatial models, where high dimensionality arises from extending the state space of the system while preserving the local stiffness and interaction range of the system.
This is in contrast to the problems with fixed-domain and refining meshes, where $\kappa$ may grow with $d$.
A representative example comparing these regimes is provided in \Cref{app:cond_scaling}.
\end{rem}

To facilitate the proof of \Cref{thm:MRF_loc}, we use \Cref{lem:diff_ctrl} below and the elementary summation bound stated in \Cref{lem:Li_ctrl}; their proofs are provided in \Cref{app:diff_ctrl,app:Li_ctrl}.
We note that \Cref{lem:diff_ctrl} controls the diffusion speed in graphs, which is crucial for establishing locality for graphical models.

\begin{lem}
\label{lem:diff_ctrl}
Let $\sfH_t \in \mR^{d\times d}$ be a time-dependent positive semi-definite matrix satisfying:
\begin{itemize}
    \item For every $t\ge 0$, $\sfH_t$ has a dependency graph $\sfG$, i.e., its $(i,j)$-th block satisfies $\sfH_{t}(i,j) = 0$ whenever $\sfd_\sfG(i,j)>1$.
    \item There exists some $M>0$ such that $0 \preceq \sfH_t \preceq MI_d$ for all $t\ge 0$.
\end{itemize}
Consider the matrix differential equation
\begin{equation}
\label{eqn:MatODE}
    \diff{}{t} \sfX_t = - \sfH_t \sfX_t, \quad \sfX_0 = I_d.
\end{equation}
Then for any $t\ge 0$, it holds that 
\begin{equation}
    \normo{ \sfX_{t}(i,j) } \le \mee^{-tM} \sum_{k=\sfd_\sfG(i,j)}^\infty \frac{t^k M^k}{k!}.
\end{equation}
\end{lem}

\subsection{Diagonally dominant probability distributions}
\label{sec:diag_dom}

Here we consider distributions whose dependency graphs are not necessarily sparse, but instead satisfy a certain diagonally dominant property.
This condition implies that each block of random variables is predominantly correlated with itself, leading to an inherent localized structure in the distribution.
Such a condition was previously studied in \cite{24M1694781} motivated by image deblurring problems, and here we show that it also implies the $\delta$-locality.

\begin{thm}
\label{thm:diag_loc}
Consider $\pi\in \mcP_1(\mR^d)$. Suppose the local precision matrix $\sfH(\bx) := - \nabla^2 \log \pi(\bx)$ is $c$-uniformly block-diagonally dominant; meaning that there exists a $c$-diagonally dominant and symmetric matrix $A \in\mR_+^{K\times K}$ with non-negative entries, i.e.,
\(
 \sum_{j\neq i} A_{ij} + c \le A_{ii},
\)
such that the blocks of $\sfH(\bx)$ satisfy
\begin{equation}
    \sfH_{ii}(\bx) \succeq A_{ii} I_{d_i} \quad \text{for} \quad i= j; \qquad \normo{ \sfH_{ij}(\bx) } \le A_{ij} \quad \text{for} \quad i\neq j,
\end{equation}
for all $i,j \in [K]$. Then, the density $\pi$ is $\delta$-localized with $\delta = c^{-1}$. 
\end{thm}

\begin{rem}
\label{rem:l_cond}
The conditions of \Cref{thm:diag_loc} imply that the density $\pi$ is log-concave. Denote $\tilde{A}_{ij}:= 2 A_{ii} \ind_{i=j} - A_{ij}$, then the matrix $\tilde{A}$ is $c$-diagonally dominant. Ger\^sgorin's discs theorem \cite{MR2978290} implies that the smallest eigenvalue of $\tilde{A}$ is lower bounded by $c$, and thus 
\begin{equation}
\begin{split}
    \bu\matT \sfH(\bx) \bu =~& \sum_{i,j} \bu_i\matT \sfH_{ij}(\bx) \bu_j \ge \sum_i A_{ii} \norm{\bu_i}^2 - \sum_{i\neq j} A_{ij} \norm{\bu_i} \norm{\bu_j} \\
    =~& \sum_{ij} \tilde{A}_{ij} \norm{\bu_i} \norm{\bu_j} \ge c \sum_{i} \norm{\bu_i}^2 = c \norm{\bu}^2,
\end{split}
\end{equation}
for all $\bu$. Therefore, $\pi$ is $c$-strongly log-concave. 
\end{rem}

The proof of \Cref{thm:diag_loc} uses an auxiliary matrix infinity-norm estimate stated in \Cref{lem:InfCtrl}.

\subsection{A Langevin semigroup perspective of \texorpdfstring{$\delta$}{delta}-locality}
\label{sec:semigroup}

In this section, we provide an alternative perspective to understand the $\delta$-locality established in \Cref{thm:MRF_loc} and \Cref{thm:diag_loc} through the lens of the Langevin semigroup.
We show that the $\delta$-locality can essentially be derived from a gradient estimate of the Langevin semigroup: 
\begin{equation}
\label{eqn:GradEstimate}
    \norm{ \nabla \sfP_t f }_{\infty,1} \lesssim \mee^{-c_\pi t} \norm{\nabla f}_{\infty,1}, 
\end{equation}
where $\norm{\cdot}_{\infty,1}$ is defined as in \eqref{eqn:PoisGradBound}, and $\sfP_t$ is the Langevin semigroup (see \cite{MR3155209}) defined as 
\begin{equation}
\label{eqn:Pt}
    \sfP_t f(\bx) = \E \Rectbrac{ f(\bX_t^\bx) },
\end{equation}
and $\bX_t^\bx$ is the solution to the (overdamped) Langevin dynamics \eqref{eqn:Langevin}.

Gradient estimates are powerful tools to derive functional inequalities and transport inequalities \cite{Guillin2009,KUWADA20103758,MR3155209}.
For instance, the gradient estimate
\begin{equation}
    \norm{ \nabla \sfP_t f (\bx) } \le \mee^{-c_\pi t}  \Rectbrac{ \sfP_t \Brac{ \norm{\nabla f}^p } (\bx) }^{1/p}.
\end{equation}
implies the Poincar\'e inequality ($p=2$) and the logarithmic Sobolev inequality ($p=1$) \cite{MR3155209}. 
In \eqref{eqn:GradEstimate}, we use the norm $\norm{\cdot}_{\infty,1}$ to adapt this idea to the locality structure, so that the gradient estimate can be used to quantify the locality of the distribution.

The $\delta$-locality condition can be directly derived from the gradient estimate \eqref{eqn:GradEstimate}.
Since $\sfL_\pi$ is the infinitesimal generator of the Langevin semigroup $\sfP_t$, formally we have 
\begin{equation}
    \sfL_\pi^{-1} = - \int_0^\infty \mee^{t\sfL_\pi} \mdd t = - \int_0^\infty \sfP_t \mdd t. 
\end{equation}
This formal identity can be made rigorous in suitable function spaces, and corresponds to \eqref{eqn:Solu} in \Cref{lem:ExpSolu}.
Let $u$ solve the marginal Stein equation \eqref{eqn:PoisEqn}. Without loss of generality, suppose that $\int u(\bx) \pi(\bx) \mdd \bx = \int \phi(\bx_i) \pi(\bx) \mdd \bx = 0$.
Denote $\psi(\bx) := \phi(\bx_i)$. Then it holds that
\(
    u = \sfL_\pi^{-1} \psi. 
\)
Therefore, if \eqref{eqn:GradEstimate} holds, then we have 
\begin{equation}
\begin{split}
    \norm{\nabla u}_{\infty,1} =~& \norm{ - \int_0^\infty \nabla \sfP_t \psi \mdd t  }_{\infty,1} \le \int_0^\infty  \norm{ \nabla \sfP_t \psi }_{\infty,1} \mdd t \\
    \lesssim~& \int_0^\infty \mee^{-c_\pi t} \norm{\nabla \psi}_{\infty,1} \mdd t \lesssim \norm{\nabla \psi}_{\infty,1} = \norm{\nabla_i \phi}_{L^\infty} \le 1.
\end{split}
\end{equation}
The hidden constant depends on $c_\pi$ and on the constant implicit in \eqref{eqn:GradEstimate}, and is typically dimension-independent.

Now we reformulate \Cref{thm:MRF_loc} and \Cref{thm:diag_loc} in terms of the Langevin semigroup. 

\begin{thm}
\label{thm:semigroup_loc}
\textup{(1)} Under conditions in \Cref{thm:MRF_loc}, it holds that 
\begin{equation}
\label{eqn:semigroup_conv_cont}
    \norm{ \nabla \sfP_t f }_{\infty,1} \le \mee^{-mt} \Rectbrac{ 1+ S p_\nu ( t(M-m) ) } \norm{\nabla f}_{\infty,1}. 
\end{equation}
Here $p_\nu(x) := \mee^{-x} \sum_{q=0}^\infty q^\nu \frac{x^q}{q!}$ is a monic polynomial of degree $\nu$.
\\
\textup{(2)} Under conditions in \Cref{thm:diag_loc}, it holds that 
\begin{equation}
    \norm{ \nabla \sfP_t f }_{\infty,1} \le \mee^{-ct} \norm{\nabla f}_{\infty,1}.
\end{equation}
\end{thm}

The proof is provided in \Cref{app:thm_Semigroup_loc}.
The constants in \Cref{thm:semigroup_loc} are dimension-independent. \Cref{thm:MRF_loc} and \Cref{thm:diag_loc} are direct consequences of \Cref{thm:semigroup_loc} using the arguments above.
The fact that $p_\nu$ is a monic polynomial of degree $\nu$ is proved in \Cref{lem:pk}.

\begin{rem}
The delocalization result of \cite{chen2026convergence} can also be interpreted from the semigroup perspective developed above.
To control the marginal bias of the unadjusted Langevin algorithm, \cite{chen2026convergence} couples its discretization with the continuous Langevin dynamics in the $\sfW_{2,\ell^\infty}$ metric.
A key step in obtaining bounds with benign dimension dependence is a multistep coupling argument: although a one-step contraction is generally unavailable in this metric, the coupling becomes contractive after a suitably chosen number of steps.
The need for this multistep analysis has a natural analogue in \Cref{thm:semigroup_loc}.
Strong convexity yields the exponential factor $\mee^{-mt}$ in \eqref{eqn:semigroup_conv_cont}, whereas the propagation of local interactions contributes the polynomial factor $1+S p_\nu(t(M-m))$.
Consequently, the bound need not be contractive for short times; contraction emerges only after the exponential decay dominates the polynomial transient factor.
From this perspective, the multistep coupling in \cite{chen2026convergence} is the discrete-time analogue of waiting until the continuous semigroup enters its contractive regime.
\end{rem}

\section{Error analysis for \texorpdfstring{$\delta$}{delta}-localized density approximations}
\label{sec:MarginTala}

Here we establish the core result of this paper.
For an arbitrary target probability density $\pi \in \mcP_{1}(\mR^d)$, \Cref{thm:margin} characterizes how accurately its marginal distributions can be approximated by the marginals of a $\delta$-localized approximation $\pi' \in \mcP_{1}(\mR^d)$.
We also extend this result to marginals over multiple blocks in \Cref{thm:margin_multi}.
These results form the foundation for our later analyses of localization in inverse problems (see \Cref{sec:LLIS}) and in score-matching methods (see \Cref{sec:LSM}).

\begin{thm}
\label{thm:margin}
Consider a target $d$-dimensional random vector $\bX$ with a density $\pi \in \mcP_1(\mR^d)$, partitioned into $K$ disjoint subsets, $\bX = (\bX_j)_{j \in [K]}$.
Suppose $\pi'\in \mcP_1(\mR^d)$ is a $\delta$-localized approximation to $\pi$ in the sense of \Cref{def:PoisGrad}.
Then, the maximal $\sfW_1$ distance between marginal distributions of $\pi,\pi'$ satisfies
\begin{equation}
\label{eqn:MarginIneqn}
    \max_{i\in [K]} \sfW_1 (\pi_i^{}, \pi_i') \le \delta \cdot \max_{j\in [K]} \norm{ \nabla_j \log \pi' - \nabla_j \log \pi }_{L^1(\pi)},
\end{equation}
where $\pi_i$ and $\pi_i'$ denote the marginal densities of $\pi$ and $\pi'$ on the block of variables $\bX_i$, respectively. 
\end{thm}

\noindent The proof is provided in \Cref{app:thm_margin}.

It is worth mentioning that marginal errors can also be controlled directly by applying the Otto--Villani inequality \cite{MR1760620} to each marginal, which gives
\begin{equation}
    \sfW_2(\pi_i^{},\pi_i') \le C_{\pi_i} \sqrt{ \E_{\pi_i}\norm{ \nabla \log \pi_i - \nabla \log \pi_i' }^2 }.
\end{equation}
However, the key limitation of this approach is that, in practice, we typically have access only to the full densities $\pi$ and $\pi'$, not to their marginals.
Computing these marginals requires integrating out all remaining components, which is often computationally prohibitive in high dimensions. As a result, the above inequality becomes difficult to apply in practical settings.
There is also a simple averaged-marginal result obtained from standard transport inequalities.
For any coupling $\gamma$ of $(\bX,\bY)$ with marginals $\pi,\pi'$,
\begin{equation}
\label{eqn:avg_margin_baseline}
    \frac{1}{K}\sum_{i=1}^K \sfW_1(\pi_i,\pi_i')
    \le \frac{1}{K}\E_\gamma \sum_{i=1}^K \norm{\bX_i-\bY_i}
    \le \frac{1}{\sqrt K}\Rectbrac{\E_\gamma \norm{\bX-\bY}^2}^{1/2}.
\end{equation}
Optimizing over couplings and applying the Otto--Villani inequality when $\pi$ is $m$-strongly log-concave gives
\begin{equation}
    \frac{1}{K}\sum_{i=1}^K \sfW_1(\pi_i,\pi_i')
    \le \frac{1}{\sqrt K} \sfW_2(\pi,\pi')
    \le \frac{1}{m\sqrt K} \Rectbrac{ \E_{\pi'} \norm{\nabla \log \pi-\nabla \log \pi'}^2}^{1/2}.
\end{equation}
Thus dimension-independent control of an averaged marginal error is already available from global transport estimates; this is related to the delocalization perspective discussed in \cite{chen2026convergence,lacker2025hierarchical}.
The contribution of \Cref{thm:margin} is the stronger worst-block control, which provides \emph{uniform} control over all marginals.

In \Cref{thm:margin}, we show that the marginal distance between a $\delta$-localized approximation $\pi'$ and the target distribution $\pi$ on a single block of variables $\bx_i$ is independent of the full dimension $d$.
We now extend this result to marginals over multiple blocks of variables. 
For a set of indices $\mcJ = \{j_1, \ldots, j_s\} \subseteq [K]$, we denote by $\bx_\mcJ$ the collection of variables corresponding to the coordinates $\cup_{j \in \mcJ} \mcI_{j}$ and denote by $d_\mcJ = |\cup_{j \in \mcJ} \mcI_{j}|$ the dimension of $\bx_\mcJ$.

\begin{thm}
\label{thm:margin_multi}
Consider two distributions $\pi, \pi' \in \mcP_1(\mR^d)$. Assume further $\pi'$ is a $\delta$-localized distribution belonging to a class that satisfies either the conditions of \Cref{thm:MRF_loc} or those of \Cref{thm:diag_loc}.
Then for any index set $\mcJ\subseteq [K]$, the $1$-Wasserstein distance between marginals of $\pi,\pi'$ over the variables $\bx_\mcJ$ satisfies 
\begin{equation}
    \sfW_1 (\pi_\mcJ^{}, \pi_\mcJ') \le \delta \; |\mcJ| \; \max_{j\in [K]} \norm{ \nabla_j \log \pi' - \nabla_j \log \pi }_{L^1(\pi)}.
\end{equation}
Here, $\delta$ is the locality parameter provided by \Cref{thm:MRF_loc} or \Cref{thm:diag_loc}, depending on which class $\pi'$ belongs to.
\end{thm}

\noindent The proof is provided in \Cref{app:thm_margin_multi}.

\Cref{thm:margin_multi} provides further control on the cross-block interactions between different blocks in $\pi'$, which cannot be directly derived from \Cref{thm:margin}.
Roughly speaking, when $\pi$ and $\pi'$ are both Gaussians, \Cref{thm:margin} only guarantees that the diagonal blocks of the covariance matrix of $\pi'$ are close to those of $\pi$, while \Cref{thm:margin_multi} further guarantees that the off-diagonal blocks are also close.

\section{Application I: localized likelihood-informed subspace}
\label{sec:LLIS}

Many high-dimensional distributions possess a hybrid low-dimensional structure that combines locality and low-rank features.
Existing dimension-reduction techniques, such as the likelihood-informed subspace (LIS) method \cite{MR3274599,MR3543164,MR4435948}, are often designed to leverage low-rank structure alone.
As a result, they may overlook the locality present in many problems, and solely rely on global low-rank features that can be computationally expensive and less physically interpretable.

In this section, we propose a \emph{localized likelihood-informed subspace} (LLIS) method by incorporating domain decomposition into the LIS framework.
This approach effectively exploits the hybrid low-dimensional structure, leading to better computational efficiency.
Using the marginal Otto--Villani inequality established in \Cref{sec:MarginTala}, we can derive dimension-independent error bounds for the marginals of the approximate distribution obtained by the localized likelihood-informed subspace method.

\subsection{Review of likelihood-informed subspace}
\label{sec:LIS_review}

The likelihood-informed subspace method is a dimension-reduction technique for efficiently sampling from a high-dimensional distribution of the form
\begin{equation}
\label{eqn:posterior}
    \pi(\bx) = \frac{1}{Z} \ell(\bx) \pi_0(\bx),
\end{equation}
where $\pi_0(\bx)$ is the prior distribution, $\ell(\bx)$ is the likelihood function, and $Z$ is the normalizing constant. 
Sampling from such distributions is a central problem in Bayesian inference, but due to the high dimensionality, it is often computationally difficult to sample directly from $\pi(\bx)$.

The likelihood-informed subspace method mitigates this challenge by exploiting low-dimensional structure in the likelihood function.
The key idea is that the likelihood function $\ell(\bx)$ often exhibits a low-rank structure, making it possible to be approximated by a ridge function
\begin{equation}
    \ell(\bx) \approx \ell_r(\bx_r), \quad \bx_r = \Pi_r \bx,
\end{equation}
where $\Pi_r$ is a projection to a low-dimensional likelihood-informed subspace $\mcX_r \subseteq \mcX$. This leads to an approximation of the target distribution as 
\begin{equation}
    \pi(\bx) \approx \pi'(\bx) \propto \ell_r(\bx_r) \pi_0(\bx).
\end{equation}
With a well-chosen subspace $\mcX_r$ and ridge approximation $\ell_r(\bx)$, the likelihood-informed subspace method provides a certified approximation \cite{MR4435948,MR4474562}, significantly reducing the sampling complexity. The method has been successfully applied to many high-dimensional Bayesian inference problems \cite{MR3274599,MR3422405}.

Identification of the subspace $\mcX_r$ is the key step in the likelihood-informed subspace method.
This subspace is designed to capture directions in which the target distribution $\pi$ differs most from the prior distribution $\pi_0$.
In \cite{MR3274599,MR4474562}, $\mcX_r$ is constructed using the principal eigenvectors (i.e., the eigenvectors of the leading eigenvalues) of the \emph{diagnostic matrix}:
\begin{equation}
    \sfQ = \int \nabla \log \ell(\bx) \nabla \log \ell(\bx) \matT \pi(\bx) \mdd \bx.
\end{equation}
By construction, the subspace spanned by the principal eigenvectors of $\sfQ$ captures the dominant variations of $\log \ell$, which guarantees that the ridge approximation $\ell_r(\bx_r)$ can accurately approximate $\ell(\bx)$.
For numerical methods to compute $\sfQ$ and other design choices of the diagnostic matrix, we refer to \cite{MR4474562} for a comprehensive review.

\subsection{Localized likelihood-informed subspace}
\label{sec:LLIS_method}

In the original likelihood-informed subspace method, the informed subspace $\mcX_r$ is constructed globally, which may not fully capture the spatial structure in the target distribution.
Moreover, highly informative parameter-to-observable maps can lead to subspaces of very high rank, which cause difficulties in algorithmic design.
A natural remedy is to exploit locality in the parameter-to-observable relationship.
In many Bayesian inference problems, the likelihood function consists of a collection of localized parameter interactions; examples include X-ray tomography, log-Gaussian Cox model used in spatial statistics, and ice-sheet dynamics driven by basal friction.
Such local, short-range interactions among parameters have been rigorously established; see, for instance, \cite{MR4744996,MR4608791} and the references therein.
To extend likelihood-informed subspace methods to localized distributions, we propose to decompose the parameters into small blocks and apply the likelihood-informed subspace method in each block to construct localized likelihood-informed subspaces.

\begin{asm}
\label{asm:LLIS}
To facilitate the construction of local likelihood-informed subspaces, we partition the parameter space $\mcX = \mR^d$ into $K$ disjoint blocks, $\bx = (\bx_j)_{j\in[K]},$ where $\bx_j \in \mR^{d_j},\, \sum_{j=1}^K d_j = d$,
and assume the target distribution $\pi$ satisfies the following conditions.
\begin{itemize}
    \item The prior $\pi_0$ and the posterior $\pi$ share the same dependency graph $\sfG$; that is, for any $i,j\in\sfV$ that are not adjacent in $\sfG$, their mixed partial derivatives vanish:
    \begin{equation}
        \nabla_{ij}^2 \log \pi_0 (\bx) = \nabla_{ij}^2 \log \pi (\bx) = 0, \quad \forall i \notin \mcN(j),~ \forall j \in \sfV.
    \end{equation}
    Moreover, $\sfG$ is $(S,\nu)$-local (see \Cref{def:loc_graph}).
    \item There exist constants $m,M > 0$ such that for all $\bx \in \mR^d$,
    \begin{equation}
        m I_d \preceq - \nabla^2 \log \pi_0(\bx) \preceq M I_d, \qquad m I_d \preceq - \nabla^2 \log \pi(\bx) \preceq M I_d.
    \end{equation}
    Here $m,M$ are the same constants as those in \Cref{thm:MRF_loc} which guarantee that the local precision matrices are well conditioned.
\end{itemize}
\end{asm}

Note that, under condition (a) of the above assumption, the likelihood function $\ell(\bx)$ is also localized in the sense that
\begin{equation}
    \nabla_{ij}^2 \log \ell(\bx) = \nabla_{ij}^2 \log \pi(\bx) - \nabla_{ij}^2 \log \pi_0(\bx) = 0, \quad \forall i \notin \mcN(j),~ \forall j \in \sfV.
\end{equation}

\subsubsection{Construction of localized likelihood-informed subspaces} 
\label{sec:LLIS_construct}

In the following, we show how to construct localized likelihood-informed subspaces when the prior is a Gaussian distribution
\(
    \pi_0 = \GN(0,\sfC),
\)
or can be transformed into a Gaussian distribution via some invertible transformation \cite{MR4500900}.
%
% For each block of variables $\bX_j$ with associated index set $\mcI_j = \{j_1,\dots,j_{d_j}\}$, we collect the corresponding canonical basis vectors into the matrix
% \begin{equation}
% \label{eqn:E_j}
%     \sfE_j = [e_{j_1}, \ldots, e_{j_{d_j}}] \in \mR^{d \times d_j},
% \end{equation}
% where $e_{i}$ denotes the $i$-th canonical basis vector of $\mR^d$. We first define
% \[
%     \mcS_j = \mathrm{range}(\sfE_j), 
% \]
% the $d_j$-dimensional subspace of $\mR^d$ in which $\bX_j$ takes its value.
%

For the $j$-th block of the variables $\bx_j \in \mR^{d_j}$, we construct its \emph{local} likelihood-informed subspace $\mcX_{j,r} \subseteq \mR^{d_j}$ by considering the following \emph{local} diagnostic matrix:
\begin{equation}
\label{eqn:diag_matrix}
    \sfQ_j = \int \Brac{ \nabla_j \log \ell(\bx) \nabla_j \log \ell(\bx) \matT + \nabla_j^{} \nabla\matT \log \ell(\bx)\, \nabla \nabla_j\matT \log \ell(\bx) } \pi(\bx) \mdd \bx.
\end{equation}
Due to the locality assumption, the second term in \eqref{eqn:diag_matrix} can be written as
\begin{equation}
    \nabla_j^{} \nabla\matT \log \ell(\bx)\, \nabla \nabla_j\matT \log \ell(\bx) = \sum_{k\in \mcN(j)} \nabla_{jk}^2 \log \ell(\bx)\, \nabla_{kj}^2 \log \ell(\bx).
\end{equation}
% when the target density is equipped with a undirected dependency graph.
Compared to the original likelihood-informed subspace method \cite{MR3274599}, which uses only first-order derivatives, our diagnostic matrix also leverages second-order information.
We also note that locality allows each $\sfQ_j$ to be computed using only local derivatives, i.e., $\nabla_j \log \ell(\bx)$ and $\nabla_{jk}^2 \log \ell(\bx)$ for $k\in\mcN(j)$, maintaining computational efficiency.

To construct the local subspace for block $\bx_j$, we consider the generalized eigenvalue problem
\begin{equation}
\label{eqn:local_gep}
    \sfQ_j^{} \bu_{\alpha} = \lambda_\alpha \sfC_j^{-1} \bu_\alpha,
\end{equation}
where $\sfC_j$ denotes the marginal prior covariance. If the prior is specified through its precision matrix $\sfC^{-1}$, then $\sfC_j$ can be computed using the Schur complement of $\sfC^{-1}$.
%
% Within $\mcS_j$, 
The local likelihood-informed subspace $\mcX_{j,r}$ is then constructed as the space spanned by the $r_j$ principal eigenvectors of the generalized eigenvalue problem.
To be specific, the local subspace is given by 
\begin{equation}
\label{eqn:local_LIS}
    \mcX_{j,r} := \text{range}(\sfU_j) \quad \text{where} \quad \sfU_j = [\bu_1,\dots, \bu_{r_j}] \in \mR^{d_j \times r_j},
\end{equation}
where $\{\bu_\alpha\}_{\alpha=1}^{r_j}$ are orthonormal eigenvectors with respect to the weighted inner product induced by $\sfC_j^{-1}$, i.e., $\sfU_j\matT \sfC_j^{-1} \sfU_j = I_{r_j}$, and $r_j = \rk (\mcX_{j,r}) $ is the number of selected directions, which can be chosen as (see \cite{MR4474562})
\begin{equation}
    r_j = \min \left\{1\le r \le d_j : \sum_{\alpha=1}^r \lambda_\alpha \ge (1-\epsilon) \tr [ \sfC_j \sfQ_j ] \right\},
\end{equation}
for some prescribed threshold $\epsilon \in (0,1)$.

We then define the joint local likelihood-informed subspace as the direct sum of local likelihood-informed subspaces:
\begin{equation}
\label{eqn:global_LLIS}
    \mcX_r = \bigoplus_{j=1}^K \mcX_{j,r}.
\end{equation}
The dimension of the informed localized subspace $\mcX_r$ is $\dim \mcX_r = \sum_{j=1}^K r_j$.
Unlike the original likelihood-informed subspace method, the dimension of the localized subspace may scale as $\mcO(d)$, and can therefore be much larger. Nevertheless, locality ensures that the computational complexity does not necessarily increase, since all computations are restricted to local blocks and can be performed in parallel, similar in spirit to the Metropolis-within-Gibbs sampler \cite{MR4111677,24M1694781}.
The larger dimension also offers the advantage that the informed subspace $\mcX_r$ can better capture physically meaningful local features, leading to more accurate approximation.

\subsubsection{Posterior approximation via marginalization}
\label{sec:LLIS_approx}

In this section, we introduce the posterior approximation induced by the localized likelihood-informed subspace.
We aim to approximate the likelihood function by averaging over the complement of the informed subspace.

We first decompose the parameter $\bx$ into the informed and uninformed components.
Equip each block $\bx_j \in \mR^{d_j}$ with the $\sfC_j^{-1}$-weighted inner product, i.e., for any $\bv,\bw \in \mR^{d_j}$, let $\ip{\bv}{\bw} = \bv\matT \sfC_j^{-1} \bw$.
Let $\Pi_{j,r} \in \mR^{d_j \times d_j}$ denote the $\sfC_j^{-1}$-orthogonal projector onto the local subspace $\mcX_{j,r}$. 
Explicitly, this projector can be expressed as
\begin{equation}
\label{eqn:local_proj}
    \Pi_{j,r} = \sfU_j^{} \sfU_j^\top \sfC_j^{-1}.
\end{equation}
Recall that $\sfU_j \in \mR^{d_j \times r_j}$ is the $\sfC_j^{-1}$-orthonormal basis of the local subspace $\mcX_{j,r}$ \eqref{eqn:local_LIS}.
For the block $\bx_j \in \mR^{d_j}$, we decompose it as
\begin{equation}
\label{eqn:block_decomp}
    \bx_{j,r} = \Pi_{j,r} \bx_j = \sfU_j^{} \sfU_j^\top \sfC_j^{-1} \bx_j , \quad \bx_{j,\bot} = ( I_{d_j} - \Pi_{j,r}) \bx_j =: \Pi_{j,\bot} \bx_j,
\end{equation}
where $\bx_{j,r} \in \mcX_{j,r}$ and $\bx_{j,\bot} \in (\mcX_{j,r})^\bot$ denote the informed and uninformed components of $\bx_j$, respectively.
To obtain the global decomposition, we define the block-diagonal projections
\begin{equation}
\label{eqn:global_proj}
    \Pi_r = \begin{bmatrix}
    \Pi_{1,r} & & & \\ & \Pi_{2,r} & & \\ & & \ddots & \\ & & & \Pi_{K,r}
    \end{bmatrix}, \quad \Pi_\bot = I_d - \Pi_r = \begin{bmatrix}
    \Pi_{1,\bot} & & & \\ & \Pi_{2,\bot} & & \\ & & \ddots & \\ & & & \Pi_{K,\bot}
    \end{bmatrix}.
\end{equation}
Then $\Pi_r$ and $\Pi_\bot$ are projections onto the global informed subspace $\mcX_r$ \eqref{eqn:global_LLIS} and its orthogonal complement $(\mcX_r)^\bot$, respectively. 
Note here the orthogonality is defined with respect to the block-diagonal prior precision matrix
\begin{equation}
    \hat{\sfC}^{-1} = \begin{bmatrix}
    \sfC_1^{-1} & & & \\ & \sfC_2^{-1} & & \\ & & \ddots & \\ & & & \sfC_{K}^{-1}
    \end{bmatrix}.
\end{equation}
Accordingly, we write 
\begin{equation}
\label{eqn:global_decomp}
    \bx_r = \Pi_r \bx = (\bx_{1,r},\cdots,\bx_{K,r}), \quad \bx_\bot = \Pi_\bot \bx = (\bx_{1,\bot},\cdots,\bx_{K,\bot}).
\end{equation}
Next, as in \cite{MR4474562}, we define the approximate likelihood $\ell_r(\bx_r)$ by averaging $\ell(\bx)$ in the logarithm regime over the uninformed subspace $(\mcX_r)^\bot$, and approximate the target distribution by
\begin{equation}
\label{eqn:LLIS_approx}
    \pi'(\bx) = \frac{1}{Z_r} \ell_r(\bx_r) \pi_0(\bx),
\end{equation}
\begin{equation}
\label{eqn:LLIS_approx2}
    \text{where} \quad \log \ell_r(\bx_r) := \int \log \ell (\bx_r,\bx_\bot) \pi_0(\bx_\bot|\bx_r) \mdd \bx_\bot,
\end{equation}
and $Z_r = \int \ell_r(\bx_r) \pi_0(\bx) \mdd \bx$ is the normalizing constant.
Here we choose to average in the logarithm regime for the sake of simplicity of the theoretical results.
In practice, one may adopt alternative averaging schemes; see \cite{MR4474562} for a detailed discussion.
Moreover, the approximate likelihood \(\ell_r(\bx_r)\) can be evaluated efficiently via Monte Carlo sampling (see \cite{MR3274599,MR4474562}).

The following theorem, proved in \Cref{app:LLIS}, shows that under \Cref{asm:LLIS} the marginal approximation error of \eqref{eqn:LLIS_approx} is independent of the ambient dimension. This error is controlled by the eigenvalue residues of the modified diagnostic matrices \eqref{eqn:diag_matrix}. 

\begin{thm}
\label{thm:LLIS_approx}
Assume that the prior distribution $\pi_0$ is Gaussian and both $\pi_0$ and $\pi$ satisfy \Cref{asm:LLIS}.
Consider the localized likelihood-informed subspace approximation $\pi'$ in \eqref{eqn:LLIS_approx}, where the informed subspace $\mcX_r$ \eqref{eqn:global_LLIS} is constructed using the local diagnostic matrices \eqref{eqn:diag_matrix}.
Then the marginal approximation error satisfies
\begin{equation}
\label{eqn:LIS_approx}
    \max_{i\in \sfV} \sfW_1(\pi_i,\pi_i') \le C \Brac{ \max_{j \in [K]} \tr \Rectbrac{ \Pi_{j,\bot} \sfR_j \Pi_{j,\bot}\matT } + \max_{j \in [K]} \tr \Rectbrac{ \Pi_\bot \widetilde{\sfR}_j \Pi_\bot\matT } }^{1/2}.
\end{equation}
Here $C$ is a dimension-independent constant that depends on $S,\nu,\kappa,m$ only, and
\begin{equation}
\label{eqn:mat_resi}
\begin{split}
    \sfR_j  & :=  \int \nabla_j \log \ell(\bx) \nabla_j \log \ell(\bx) \matT \pi'(\bx) \mdd \bx \in \mR^{d_j \times d_j},  \\
    \widetilde{\sfR}_j & := \int \nabla \nabla_j\matT \log \ell(\bx)\, \nabla_j^{} \nabla\matT \log \ell(\bx)\, \pi'(\bx) \mdd \bx \in \mR^{d \times d}.
\end{split}
\end{equation}
\end{thm}

\begin{rem}
Note that $ \widetilde{\sfR}_j (i,k) = 0$ if $i\notin \mcN(j)$ or $k\notin \mcN(j)$, so that only a few blocks in the matrix $\widetilde{\sfR}_j$ are nonzero, i.e.,
\begin{equation}
    \widetilde{\sfR}_j(i,k) = \ind_{i\in \mcN(j)} \ind_{k\in \mcN(j)} \int \nabla_{ij}^2 \log \ell(\bx)\, \nabla_{jk}^2 \log \ell(\bx)\, \pi'(\bx) \mdd \bx.
\end{equation}
Thus the second term in \eqref{eqn:LIS_approx} is dimension-independent.
\end{rem}

\begin{rem}
The above theorem quantifies the approximation error of $\pi'$ using \emph{arbitrary} local projections of the form \eqref{eqn:global_proj}.
Ideally, one should choose the local projections $\Pi_{j,r}$ to minimize the approximation error bound \eqref{eqn:LIS_approx}.
Since $\sfR_j$ and $\widetilde{\sfR}_j$ are not directly computable because they depend on the approximate distribution $\pi'$, we can instead replace $\pi'$ by $\pi$ and use the diagnostic matrices $\sfQ_j$ defined in \eqref{eqn:diag_matrix} to construct the local projections $\Pi_{j,r}$.
In general, one can also consider other constructions of the local subspaces and projections to minimize the error bound in \eqref{eqn:LIS_approx}.
\end{rem}

\section{Application II: localized score matching}  \label{sec:LSM}

Score matching is a variational approach for approximating a probability distribution $\pi \in \mcP_{1}(\mR^d)$ by fitting its score function $s(\bx) := \nabla \log \pi(\bx)$.
Originally introduced in \cite{JMLR:v6:hyvarinen05a} and later extended in \cite{NEURIPS2019_3001ef25} to an important denoising variant, score matching estimates a target distribution using only data $(\bX^{(1)},\dots,\bX^{(N)})$ sampled from it and is commonly used to train generative models.

A central concern in this context is the sample efficiency of score-based density estimation.
As shown in \cite{pmlr-v202-oko23a,pmlr-v247-wibisono24a}, the overall score matching error scales as $\mcO(N^{-\gamma/(d+2\gamma)})$, where $\gamma > 0$ characterizes the smoothness of the target density.
Thus, the number of samples required to achieve a fixed accuracy $\varepsilon$ scales as $N = \mcO(\varepsilon^{-(d+2\gamma)/\gamma})$, which grows rapidly with $d$.
Our objective in this section is to show how score matching can exploit an a priori local graphical structure.
The reduction in statistical complexity comes from the local factorization of the target distribution, as in other procedures for sparse Markov random fields and factor graphs; score matching provides a convenient sample-based objective for fitting the resulting local score components.

\subsection{Review of score matching}
\label{sec:SM_review}

Given a parametric family of distributions $\{\pi_\theta:\theta \in\Theta\}$, the score matching method aims to find the parameter $\theta$ that minimizes the Fisher divergence
\begin{equation}
    \sfI(\pi\|\pi_\theta) := \E_{\bx \sim \pi} \norm{ s_\theta(\bx) - \nabla \log\pi (\bx) }^2, \quad s_\theta(\bx) := \nabla \log \pi_\theta(\bx).
\end{equation}
Because the score function does not depend on the normalizing constant, minimizing the Fisher divergence is advantageous when dealing with unnormalized densities.
Nonetheless, the Fisher divergence cannot usually be evaluated explicitly, as the target score function $\nabla \log\pi(\bx)$ is unknown and only samples from the target distribution $\pi$ are available in many practical scenarios.
Under suitable regularity conditions, integration by parts yields the tractable objective
\begin{equation}
\label{eqn:SM_loss}
    J(\theta) := \E_{\bx \sim \pi} \Rectbrac{ 2 \tr \Brac{ \nabla s_\theta(\bx) } + \norm{s_\theta(\bx)}^2 },
\end{equation}
under the usual vanishing boundary conditions on $\pi$ and $s_\theta$.
Note that $J(\theta)$ differs from the Fisher divergence $\sfI(\pi\|\pi_\theta)$ only by a constant,
\begin{equation}
\label{eqn:SM_equiv}
    J(\theta) = \sfI(\pi\|\pi_\theta) - \E_{\bx \sim \pi} \norm{ \nabla \log \pi(\bx) }^2.
\end{equation}
Thus, minimizing the objective function $J(\theta)$ is equivalent to minimizing the Fisher divergence.
Given data $(\bX^{(1)},\dots,\bX^{(N)})$ sampled from the target $\pi$, we estimate $J(\theta)$ by
\begin{equation}
    \widehat{J}(\theta) = \frac{1}{N} \sum_{i=1}^N \Rectbrac{ 2 \tr\, (\nabla s_\theta(\bX^{(i)})) +  \normo{ s_\theta(\bX^{(i)})}^2 }. 
\end{equation}
Many score matching methods parameterize the score function $s_\theta(\bx)$ and minimize $\widehat{J}(\theta)$ to fit it \cite{JMLR:v6:hyvarinen05a,NEURIPS2019_3001ef25}.
Here we consider directly the density estimation problem \cite{NIPS2015_83adc922,JMLR:v18:16-011}, where we parameterize the (unnormalized) density function $\pi_\theta(\bx)$ itself, and optimize $\widehat{J}(\theta)$ over a hypothesis class $\msP$ of probability densities.
The hypothesis class $\msP$ is chosen so that both the score function $s_\theta(\bx)$ and its Jacobian are computationally tractable.

\subsection{Localized score matching}
\label{sec:LSM_method}

Despite its simple implementation, the sample complexity of score matching can grow rapidly with dimensionality. 
To mitigate this issue, we exploit the locality structure of the target distribution.
When the distribution can be represented, or well approximated, by a sparse graphical model, its score function inherits low intrinsic complexity.
By leveraging this property, we design a \emph{localized score matching} approach that reduces the effective dimension of the estimation problem while preserving control of the marginal approximation error.

\subsubsection{Clique factorization of sparse graphical models}
\label{sec:LSM_clique}

Localized score matching is motivated by the clique factorization of sparse graphical models \cite{clifford1990markov,MR2778120}. We assume the target distribution $\pi(\bx)$ is equipped with a dependency graph $\sfG=(\sfV,\sfE)$. Then the Hammersley--Clifford theorem \cite{clifford1971markov} gives the factorization
\begin{equation}
\label{eqn:clique_decomp}
    \log \pi(\bx) = \sum_{j\in \sfV} u_j(\bx_{\mcN(j)}). 
\end{equation}
Here $u_j$ is a function of $\bx_{\mcN(j)} \in \mR^{d_{\mcN(j)}}$, and $d_{\mcN(j)} := \sum_{k\in \mcN(j)} d_k$ is the local dimension.
See \Cref{sec:Intro_GM} for details of the block decomposition and graph notation.
We use this neighborhood-based representation as a convenient relaxation; for the exact clique formulation of the Hammersley--Clifford theorem, we refer to \cite{clifford1971markov,MR2778120}.
If $\sfG$ is $(S,\nu)$-local in the sense of \Cref{def:loc_graph}, then the \emph{maximal local dimension} satisfies
\begin{equation}
\label{eqn:loc_dim}
    d_{\rm loc} := \max_{j \in \sfV} d_{\mcN(j)} \le (1+S) \max_{j \in \sfV} d_j. 
\end{equation}
When $d_{\rm loc} \ll d$, the complexity of learning the score function $\nabla \log\pi(\bx)$ can be substantially reduced.
This motivates using a hypothesis class $\msP_\sfG$ that embeds the locality structure. We define
\begin{equation}
\label{eqn:G_hypo_class}
    \msP_\sfG = \Big\{ \pi_\theta(\bx) = \exp \Big( \sum_{j\in \sfV} u_{\theta,j}(\bx_{\mcN(j)}) \Big) \in \mcP_1(\mR^d) : u_{\theta,j} \in \mcU_j \Big\},
\end{equation}
where each $\mcU_j$ is a class of graph-sparse functions on $\mR^{d_{\mcN(j)}}$ to be specified later.
Thus any distribution in $\msP_\sfG$ has a dependency graph compatible with $\sfG$ by construction.
The metric entropy of this hypothesis space \cite{MR1385671} is controlled by the entropies of the local function classes and therefore depends on the maximal local dimension $d_{\rm loc}$ rather than directly on $d$; this dependence is quantified in \Cref{sec:LSM_samplecomplexity}.
Thus, apart from the number of local components, both learning and optimization are governed by problems of local dimension.

In this paper, we treat the graph $\sfG$ as known and specified \emph{a priori} based on domain knowledge.
Learning $\sfG$ from samples has been well studied for Gaussian graphical models \cite{MR2278363}, Ising models \cite{MR2662343}, and more general Markov random fields \cite{MR3037003,MR3734242,NIPS2017_ea8fcd92,kwon2026nonparametric}; we do not pursue this task here.
There are also methods that simultaneously estimate the graph structure and the parameters of the distribution; see, for example, \cite{MR3388257,MR3486418,MR3734242}.
These methods, however, are typically tailored to specific parametric families, such as Ising models and finite-dimensional exponential families.
The canonical-factor estimator of \cite{MR2274423} applies more broadly to discrete factor-graph distributions, but does not cover continuous distributions or nonparametric density estimation.
The only exception is a very recent work \cite{kwon2026nonparametric}, which leverages score functions of generative diffusion models to infer the underlying graphical structures of nonparametric densities. However, it assumes a fixed model dimension, and the scalability of the method remains unclear.
%A classical alternative method is the pseudolikelihood method \cite{Besag1975Pseudo}, which replaces the joint likelihood with a product of componentwise conditional densities and can likewise be localized according to the prescribed graph.
%Pseudolikelihood is also closely related to score matching; see \cite{JMLR:v6:hyvarinen05a,DBLP:conf/iclr/KoehlerHR23}.
%The localized score matching can generalize the localization principle to nonparametric density estimation for continuous Markov random fields with a prescribed graph.
%
A classical alternative is the pseudolikelihood method \cite{Besag1975Pseudo}, which is also closely related to score matching \cite{JMLR:v6:hyvarinen05a,DBLP:conf/iclr/KoehlerHR23}. It replaces the joint likelihood with a product of componentwise conditional densities, and hence this approach could potentially benefit from the localization strategy discussed here. 
In particular, our localized score matching extends the componentwise localization of pseudolikelihood to nonparametric density estimation for continuous Markov random fields with a prescribed graph. 
By decomposing the global problem into low-dimensional local score-estimation problems, the sample complexity is expected to depend mainly on the local dimension rather than the ambient dimension. Moreover, we establish uniform componentwise control of the score-estimation error, which in turn yields accurate approximations of low-dimensional marginals of the target distribution.

\subsubsection{Localized score matching}
\label{sec:LSM_detail}

We now introduce the localized score matching method, which exploits the clique factorization \eqref{eqn:clique_decomp} to reduce the sample complexity of density estimation.
By the clique factorization \eqref{eqn:clique_decomp}, the $j$-th component of the score function of $\pi$ is
\begin{equation}
    s_j(\bx) := \nabla_j \log \pi(\bx) = \sum_{k \in \mcN(j)} \nabla_j u_k(\bx_{\mcN(k)}),
\end{equation}
which we refer to as the \emph{local score function}. To estimate the local score function, it suffices to minimize
\(
    \E_{\bx \sim \pi} \norm{ s_{\theta,j}(\bx) - s_j(\bx) }^2,
\)
where $s_{\theta,j}(\bx) := \nabla_j \log \pi_\theta (\bx)$ is the $j$-th localized approximation. By applying integration by parts, this becomes equivalent to minimizing
\begin{equation}
\label{eqn:J_j}
    J_j(\theta) := \E_{\bx \sim \pi} \Rectbrac{ 2 \tr \Brac{ \nabla_j s_{\theta,j} (\bx) } + \norm{s_{\theta,j}(\bx)}^2 }.
\end{equation}
To estimate the full score function, one can consider the min-max problem:
\begin{equation}
\label{eqn:J_minmax}
    \inf_{\theta\in\Theta} \max_{j \in \sfV}\, ( J_j(\theta) - \tau_j), \quad \text{where} \quad \tau_j := \inf_{\theta \in \Theta} J_j(\theta).
\end{equation}
Here, $\Theta$ denotes the parameter space of the hypothesis class.
We subtract the constant $\tau_j$ to ensure that, when the target distribution $\pi$ is contained in the hypothesis class, the above min-max problem is equivalent to minimizing the maximal local score error, i.e.,
\(
    \max_{j\in \sfV} \E_{\bx \sim \pi} \norm{ s_{\theta,j}(\bx) - s_j(\bx) }^2 = \max_{j\in \sfV}\, ( J_j(\theta) - \tau_j ).
\)
Since the constant $\tau_j$ is unknown, \eqref{eqn:J_minmax} is not directly solvable.
In practice, both $J_j(\theta)$ and $\tau_j$ can be replaced by their empirical counterparts.
Alternatively, one may relax \eqref{eqn:J_minmax} by minimizing the average local loss $\frac{1}{|\sfV|}\sum_{j\in \sfV} J_j(\theta)$, which has the same minimizers as the standard score matching loss $J(\theta)$ in \eqref{eqn:SM_loss}.

In practice, given the data $(\bX^{(1)},\dots,\bX^{(N)})$ sampled from $\pi$, the empirical loss corresponding to $J_j(\theta)$ defined in \eqref{eqn:J_j} is given by
\begin{equation}
\label{eqn:J_j_emp}
    \widehat{J}_j(\theta) = \frac{1}{N} \sum_{i=1}^N \Rectbrac{ 2 \tr ( \nabla_j s_{\theta,j} (\bX^{(i)}) ) + \normo{s_{\theta,j}(\bX^{(i)})}^2  }.
\end{equation} 
When $\pi_\theta$ is chosen in the localized hypothesis class $\msP_\sfG$ \eqref{eqn:G_hypo_class}, the loss $\widehat{J}_j$ can be computed locally, since the local score
\(
    s_{\theta,j}(\bx) = \sum_{k \in \mcN(j)} \nabla_j u_{\theta,k}(\bx_{\mcN(k)})
\)
depends only on the potential functions indexed by $\mcN(j)$, while the graph-sparsity condition imposed on the local potentials ensures that $s_{\theta,j}$ depends only on $\bx_{\mcN(j)}$. This enables local evaluation and optimization of the score-matching objective.

% \todo{multiple $\lambda$? why not just solve them all together? minmax is veru hard. what is the lower bound?}

\subsection{Sample complexity analysis}
\label{sec:LSM_samplecomplexity}

We now show that the sample complexity of localized score matching is effectively independent of the ambient dimension $d$.
Motivated by the sample complexity analysis in \cite{pmlr-v202-oko23a,pmlr-v247-wibisono24a}, we localize the empirical score matching loss to a compact domain while keeping the target and model densities defined on $\mR^d$.
For a truncation level $\Lambda\ge1$, define the local truncation set
\(
    \Omega_j^\Lambda := \left\{ \bx\in\mR^d : \bx_{\mcN(j)}\in[-\Lambda,\Lambda]^{d_{\mcN(j)}} \right\},
\)
and replace the empirical loss $\widehat{J}_j(\theta)$ in \eqref{eqn:J_j_emp} with
\begin{equation}
\label{eqn:J_j_trunc}
    \widehat{J}_j^\Lambda(\theta)
    =
    \frac{1}{N} \sum_{i=1}^N \Rectbrac{ 2 \tr ( \nabla_j s_{\theta,j} (\bX^{(i)}) ) + \normo{s_{\theta,j}(\bX^{(i)})}^2  }
    \ind_{\Omega_j^\Lambda}(\bX^{(i)}).
\end{equation}
Here $\ind_\Omega$ denotes the indicator of the set $\Omega$.
Our analysis (cf. \Cref{rem:truncation}) offers guidance on choosing the truncation level $\Lambda$ as a function of the sample size $N$ to ensure sufficient sample coverage, while maintaining $\min_{j\in\sfV} \pi(\Omega_j^\Lambda)$ sufficiently close to $1$ so that the truncation error remains small.

Suppose an optimization algorithm solves the truncated empirical counterpart of \eqref{eqn:J_minmax} with accuracy $\varepsilon_{\rm opt}$, i.e., it returns a parameter $\widehat{\theta}_*$ such that
\begin{equation}
\label{eqn:theta_emp}
   \widehat{J}_{\rm loc}^\Lambda (\widehat{\theta}_*) - \inf_{\theta\in\Theta} \widehat{J}_{\rm loc}^\Lambda(\theta) \le \varepsilon_{\rm opt},
\end{equation}
\begin{equation}
\label{eqn:J_loc_emp}
    \text{where} \quad \widehat{J}_{\rm loc}^\Lambda(\theta) := \max_{j\in \sfV}\, ( \widehat{J}_j^\Lambda(\theta) - \widehat{\tau}_j^\Lambda), \quad \widehat{\tau}_j^\Lambda := \inf_{\theta\in\Theta} \widehat{J}_j^\Lambda(\theta).
\end{equation}
Our goal is to quantify the convergence rate of the learned distribution $\widehat{\pi} := \pi_{\widehat{\theta}_*}$ to the target distribution $\pi$ in terms of their marginal $1$-Wasserstein distance, i.e., $\max_{i\in \sfV} \sfW_1(\pi_i,\widehat{\pi}_i)$.
A key point of our analysis is that the sample complexity remains dimension-independent up to logarithmic factors, which effectively removes the curse of dimensionality.

Given a dependency graph $\sfG=(\sfV,\sfE)$, we consider the hypothesis class
\begin{equation}
\label{eqn:hypo_class}
    \msP = \Big\{ \pi_\theta(\bx) = \exp \Big( \sum_{j\in \sfV} u_{\theta,j}(\bx_{\mcN(j)}) \Big) \in \mcP_1(\mR^d) : mI \preceq - \nabla^2 \log \pi_\theta \preceq MI,~u_{\theta,j} \in \mcU_j \Big\}.
\end{equation}
\begin{equation}
\label{eqn:U_j}
    \text{where} \quad \mcU_j = \left\{ u_{\theta,j} \in C^2(\mR^{d_{\mcN(j)}}) :
    \normo{u_{\theta,j}|_{[-\Lambda,\Lambda]^{d_{\mcN(j)}}}}_{C^\gamma} \le R\,;\,
    \partial_{ik}^2u_{\theta,j}=0~\text{for all }i\notin\mcN(k) \right\}.
\end{equation}
Here $0< m\le M< \infty$ and $R>0$ are constants. The sparsity condition on $u_{\theta,j}$ ensures that every density in $\msP$ has a dependency graph compatible with $\sfG$.
We only require that the restriction of $u_{\theta,j}$ on $[-\Lambda,\Lambda]^{d_{\mcN(j)}}$ is bounded in the $C^\gamma$ norm, where $\gamma\in \mZ_+$ and $\gamma > 2$.
We note that the requirement $\gamma \in \mZ_+$ can be relaxed to non-integer values using H\"older spaces.
The global $C^2$ regularity ensures that $\pi_\theta$ and its score remain well defined on $\mR^d$.
Because the empirical criterion constrains $u_{\theta,j}$ only on the truncation cube, we equip the restriction $u_{\theta,j}|_{[-\Lambda,\Lambda]^{d_{\mcN(j)}}}$ with a global $C^2$ convex extension that preserves the Hessian bounds in \eqref{eqn:hypo_class}.
In practice, numerical methods such as the input convex neural network \cite{pmlr-v70-amos17b} can be used to ensure the log-concavity of $\msP$. Furthermore, we do not consider the hypothesis error here for simplicity. That is, we assume that the target distribution $\pi$ belongs to $\msP$, and hence there exists $\theta^* \in \Theta$ such that $\pi = \pi_{\theta^*}$.

\begin{thm}
\label{thm:LSM}
Let $\sfG = (\sfV,\sfE)$ be a $(S,\nu)$-local graph. Suppose the target distribution $\pi(\bx)$ belongs to the hypothesis class $\msP$ defined in \eqref{eqn:hypo_class}.
Let $\widehat{\theta}_*$ be the empirical solution obtained by localized score matching with $N$ independent and identically distributed samples and optimization error $\varepsilon_{\rm opt}$ as in \eqref{eqn:theta_emp}, and denote the learned distribution by $\widehat{\pi} := \pi_{\widehat{\theta}_*}$.
Then for any $\beta\in (0,1)$, with probability at least $1-\beta$, we have 
\begin{equation}
\label{eqn:LSM_conv}
    \max_{i \in \sfV} \sfW_1(\pi_i,\widehat{\pi}_i) \lesssim \max \left\{ \left(\frac{\log(d/\beta)}{N}\right)^{1/4}, \left(\frac{\Lambda^{d_{\rm loc}}}{N}\right)^{(\gamma-2)/(2(d_{\rm loc} + 2(\gamma-2)))} \right\} + \mee^{-m\Lambda^2/16} + \varepsilon_{\rm opt}^{1/2}.
\end{equation}
Here the hidden constant in the above bound depends only on dimension-free constants $S,\nu,m,M$, $R,\gamma$, and on the local dimension $d_{\rm loc}$ \eqref{eqn:loc_dim}, but is independent of the total dimension $d$.
\end{thm}

The proof of \Cref{thm:LSM} is provided in \Cref{app:LSM}.
\begin{rem}
\label{rem:truncation}
We can take $\Lambda = \sqrt{ C \log N }$ with sufficiently large $C>0$ to ensure that the truncation term $\mee^{-m\Lambda^2/16} = N^{-Cm/16} \ll N^{-1/4}$; this choice contributes only a polylogarithmic factor in $N$ to the statistical error.
Considering the sample complexity terms in \eqref{eqn:LSM_conv}, we observe that for sufficiently large $N$, the dimension-dependent factor $\big(\log(d/\beta)/N\big)^{1/4}$ is dominated by the term
\(
\left(\Lambda^{d_{\rm loc}}/N\right)^{(\gamma-2)/(2(d_{\rm loc}+2(\gamma-2)))},
\)
which is independent of the ambient dimension $d$. Consequently, localized score matching attains dimension-independent sample complexity
\(
    N=\mcO \! \big( \varepsilon^{-\,2(d_{\rm loc}+2(\gamma-2))/(\gamma-2)} \big),
\)
up to logarithmic factors in $\varepsilon$ and $d/\beta$.
In contrast, standard score matching requires
\(
    N=\mcO\! \big( \varepsilon^{-(d+2\gamma)/\gamma} \big),
\)
which grows exponentially with $d$.
Thus, learning within a localized hypothesis class avoids the polynomial dependence of the convergence exponent on the ambient dimension.
\end{rem}

\begin{rem}
In \eqref{eqn:LSM_conv}, the convergence rate $ \frac{1}{2} \frac{\gamma-2}{d_{\rm loc} + 2(\gamma-2)} $ is not the minimax rate for estimating the target distribution $\pi$ with local dimension $d_{\rm loc}$ and smoothness $\gamma$.
First, the loss \eqref{eqn:SM_loss} involves second derivatives of the log-density, which reduces the effective smoothness from $\gamma$ to $\gamma-2$ in the metric-entropy bound.
Second, the factor $\frac{1}{2}$ arises when the squared local score error is converted to $1$-Wasserstein distance through the marginal transport inequality.
Since achieving the minimax rate is not our focus here, we leave it as future work.
\end{rem}

\section{Conclusion}
\label{sec:con}

In this paper, we consider the locality structure in high-dimensional distributions.
We introduce a quantitative analysis by proposing the $\delta$-locality condition and establishing a marginal Otto--Villani inequality. 
We develop Stein's method for analyzing the marginals, and give a new characterization of the locality structure based on Stein's method. 
The marginal form of the Otto--Villani inequality provides refined approximation guarantees for low-dimensional marginals, which lays the theoretical foundation for localization methods. 
We also propose applying the localization methodology to distribution approximation and sampling to design scalable algorithms in high dimensions.
As examples, we show how to localize the likelihood-informed subspace method and score matching to design scalable algorithms for high-dimensional distributions.

It would be interesting to extend the analysis framework to more general distributions and different locality conditions. Relaxing the log-concavity assumption and finding operations that preserve locality would be practically interesting. It is also interesting to investigate the localization methodology in other applications, such as kernel methods and modern generative models. We leave these as future work.

\section*{Acknowledgements}

The authors thank the anonymous referee for drawing their attention to the related literature on functional inequalities and graphical model learning.

The work of TC is partially supported by the Australian Research Council grant FT250100199. The work of SL is partially supported by the Overseas Research Immersion Award (ORIA) of the National University of Singapore and Singapore MOE grant Tier-1-A-8002956-00-00. 
The work of XT is partially supported by the Singapore MOE grant Tier-1-A-8002956-00-00.

\appendix

\renewcommand{\thesection}{S\arabic{section}}
\renewcommand{\thesubsection}{\thesection.\arabic{subsection}}
\renewcommand{\thesubsubsection}{\thesubsection.\arabic{subsubsection}}
\numberwithin{equation}{section}
\numberwithin{thm}{section}
\numberwithin{lem}{section}
\numberwithin{prop}{section}
\numberwithin{cor}{section}
\numberwithin{defn}{section}
\numberwithin{asm}{section}
\numberwithin{exmp}{section}
\numberwithin{rem}{section}

\section{Additional remarks}
\label{app:add_remarks}

\subsection{Condition-number scaling in spatial models}
\label{app:cond_scaling}

We give a representative example illustrating why the condition number $\kappa$ may remain bounded or grow with the dimension, depending on the high-dimensional regime.
Consider the quadratic nearest-neighbor interaction model on a one-dimensional lattice with $D$ sites:
\begin{equation}
    p(x) \propto \exp \Brac{ - \frac{1}{2} \sum_{i=1}^D x_i^2 - \frac{\gamma}{2} \sum_{i=1}^{D-1} \Brac{ \frac{x_{i+1}-x_i}{h} }^2 },
\end{equation}
where $h>0$ is the mesh size and $\gamma>0$ is the interaction strength.
Then
\begin{equation}
    - \nabla^2 \log p(x)
    =
    I + \frac{\gamma}{h^2}
    \begin{bmatrix}
        1 & -1 & 0 & \cdots & 0 \\
        -1 & 2 & -1 & \cdots & 0 \\
        0 & -1 & 2 & \cdots & 0 \\
        \vdots & \vdots & \vdots & \ddots & \vdots \\
        0 & 0 & 0 & \cdots & 1
    \end{bmatrix}.
\end{equation}
The eigenvalues are
\(
    \lambda_k = 1 + 4\gamma h^{-2} \sin^2 \frac{(k-1)\pi}{2D}
\)
for $k=1,\ldots,D$.
Hence $\lambda_{\min}=1$ and $\lambda_{\max}\asymp 1+4\gamma h^{-2}$ for large $D$.
Different scalings of $h$ with $D$ lead to different behaviors of $\kappa=\lambda_{\max}/\lambda_{\min}$.

If $x\in\mR^D$ discretizes a field on a fixed spatial domain $[0,1]$, then $h=(D-1)^{-1}$.
As $D$ increases, the mesh is refined on the same domain, and
\begin{equation}
    \kappa \asymp 1 + 4\gamma h^{-2} \asymp 4\gamma D^2,
    \qquad D\to\infty.
\end{equation}
This is a fixed-domain refinement regime in which $\kappa$ grows with the dimension.
In contrast, if $x\in\mR^D$ discretizes a field on an expanding domain $[0,(D-1)h_0]$ with fixed mesh size $h_0>0$, then
\begin{equation}
    \kappa \asymp 1 + 4\gamma h_0^{-2},
    \qquad D\to\infty.
\end{equation}
Here the number of sites grows while the local interaction scale is fixed, and the condition number remains bounded uniformly in $D$.
This expanding-domain regime is representative of the locality setting in which \Cref{thm:MRF_loc} yields a dimension-independent locality constant.

\section{Proofs of the main results}
\label{app:main_results}

\subsection{Proof of Lemma~1}
\label{app:lem_ExpSolu}

\begin{proof}[Proof of Lemma~1]
Let $\bX_t^\bx$ follow the Langevin dynamics \eqref{eqn:Langevin}. By Dynkin's formula \cite{MR2001996} and the marginal Stein equation \eqref{eqn:PoisEqn}, we have
\begin{equation}
\label{eqn:Dynkin}
\begin{split}
    \E \big[ u(\bX_T^\bx) \big] - u(\bx) =~& \E \Rectbrac{ \int_0^T \Brac{ \nabla \log \pi (\bX_t^\bx) \cdot \nabla u(\bX_t^\bx) + \Delta u(\bX_t^\bx) } \mdd t } \\
    =~& \int_0^T \E \Rectbrac{ \phi(\bX_{t,i}^\bx) - \E_\pi[\phi(\bx_i)] } \mdd t.
\end{split}
\end{equation}
When the distribution is strongly log-concave, it is well known that the law of $\bX_t^\bx$ converges to the equilibrium density $\pi$ at an exponential rate \cite{chewi2023logconcave}.
This ensures that the limit $T\gtoinf$ exists for both sides in \eqref{eqn:Dynkin}, and the limit is 
\begin{equation}
    \int u(\bx) \pi(\bx) \mdd \bx - u(\bx) = \int_0^\infty \E \Rectbrac{ \phi(\bX_{t,i}^\bx) - \E_\pi[\phi(\bx_i)] } \mdd t.
\end{equation}
This gives the solution to the marginal Stein equation.
Assume for the moment that $\phi \in C^1$.
Taking the derivative with respect to $\bx_j$ gives
\begin{equation}
\label{eqn:pf_grad_est}
    \nabla_j u(\bx) = - \int_0^\infty \E \Rectbrac{ (\nabla_{\bx_j} \bX_{t,i}^\bx)^\top \nabla \phi (\bX_{t,i}^\bx) } \mdd t. 
\end{equation}
Note that taking derivatives on both sides is valid due to the exponential decay of $\nabla_{\bx_j} \bX_{t,i}^\bx$.
Let \(\{\phi^\varepsilon\}_{\varepsilon>0}\subseteq C^1(\mR^{d_i})\) be a mollification of \(\phi\) such that $\phi^\varepsilon \to \phi$ pointwise and $\norm{\nabla \phi^\varepsilon}_{L^\infty} \le 1$.
Let \(u^\varepsilon\) be the solution to the marginal Stein equation with test function \(\phi^\varepsilon\).
Taking derivative with respect to \(\bx_j\) in \eqref{eqn:pf_grad_est}, we have
\begin{equation}
\label{eqn:mollified_grad_bound}
    \norm{\nabla_{\bx_j} u^\varepsilon(\bx)}
    \le
    \int_0^\infty
    \E\left[
        \norm{\nabla_{\bx_j} \bX_{t,i}^{\bx}}\,
        \norm{\nabla \phi^\varepsilon(\bX_{t,i}^{\bx})}
    \right]
    \mdd t
    \le
    \int_0^\infty
    \E\left[
        \norm{\nabla_{\bx_j} \bX_{t,i}^{\bx}}
    \right]
    \mdd t .
\end{equation}
Since the right-hand side is independent of \(\varepsilon\), we may pass to the limit \(\varepsilon\to0\).
Since \(u^\varepsilon\) converges to the solution \(u\) associated with \(\phi\) in $H_0^1(\pi)$, the same bound holds for \(u\). This completes the proof.
\end{proof}

\subsection{Proof of Theorem~1}
\label{app:thm_MRF_loc}

\begin{proof}[Proof of Theorem~1]

\Cref{lem:ExpSolu} holds under the conditions of \Cref{thm:MRF_loc}. Following the gradient estimate in \eqref{eqn:GradCtrl}, the remaining task is to control $\nabla_{\bx_j} \bX_{t,i}^\bx$.
Taking derivative with respect to $\bx$ in \eqref{eqn:Langevin}, we get
\begin{equation}
\label{eqn:pf_dXEvol}
    \mdd \nabla \bX_t^\bx = - \sfH_t  \nabla \bX_t^\bx \mdd t, \quad \sfH_t := \sfH(\bX_t^\bx) = - \nabla^2 \log \pi(\bX_t^\bx),
\end{equation}
where $\sfH_t$ contains partial derivatives with respect to  $\bX_t^\bx$ and $\nabla \bX_t^\bx := \nabla_\bx \bX_t^\bx$.
Denote $\sfX_t = \mee^{mt} \nabla \bX_t^\bx$ and $\widetilde{\sfH}_t = \sfH_t - m I_d $, then it holds that
\begin{equation}
    \diff{}{t} \sfX_t = \mee^{mt} \Brac{ m \nabla \bX_t^\bx - \sfH_t \nabla \bX_t^\bx } = - \widetilde{\sfH}_t \sfX_t, \quad \sfX_0 = \nabla \bX_0^\bx = I_d.
\end{equation}
Since the matrix $\sfH(\bx)$ satisfies $m I_d \preceq \sfH(\bx) \preceq M I_d$ for all $\bx$ by assumption and $\sfH_{ij}(\bx) = 0$ whenever $\sfd_{\sfG}(i,j)>1$, it follows that $0 \preceq \widetilde{\sfH}_{t} \preceq (M-m)I_d$ for all paths of the Langevin dynamics \eqref{eqn:Langevin} and $\widetilde{\sfH}_{t}(i,j) = 0$ whenever $\sfd_{\sfG}(i,j)>1$.
By \Cref{lem:diff_ctrl}, we obtain
\begin{equation}
\label{eqn:pf_dXGrad}
    \mee^{mt} \normo{ \nabla_{\bx_j} \bX_{t,i}^\bx } = \normo{ \sfX_{t}(i,j) } \le \mee^{-(M-m) t} \sum_{k=\sfd_\sfG(i,j)}^\infty \frac{t^k(M-m)^k}{k!},
\end{equation}
for any initial condition $\bx$. For every vertex $j\in \sfV$, let $\bx^{(j)}$ denote the initial condition at which the supremum over the $j$th block is evaluated. This notation is used to control $\sum_j \sup_{\bx}\|\nabla_j u(\bx)\|$, rather than $\sup_{\bx}\sum_j \|\nabla_j u(\bx)\|$. Taking summation over $j\in \sfV$, we obtain
\begin{equation}
\begin{split}
    \sum_{j\in \sfV} \normo{ \nabla_{\bx_j} \bX_{t,i}^{\bx^{(j)}} } \le~& \mee^{-mt} \mee^{-(M-m) t} \sum_{j\in \sfV} \sum_{ k = \sfd_\sfG(i,j) }^\infty \frac{t^k(M-m)^k}{k!} \\
    =~& \mee^{-mt} \mee^{-(M-m) t} \Brac{ \sum_{ k = 0 }^\infty \frac{t^k(M-m)^k}{k!} + \sum_{j\neq i} \sum_{ k = \sfd_\sfG(i,j) }^\infty \frac{t^k(M-m)^k}{k!} }.
\end{split}
\end{equation}
Using the graph-locality condition $\sum_{j \neq i} \ind_{ \sfd_\sfG(i,j) \le k }  = |\mcN'(i;k)| \le S k^\nu $ specified in \eqref{eqn:loc_graph}, we have
\begin{equation}
    \sum_{j\neq i} \sum_{ k = \sfd_\sfG(i,j) }^\infty \frac{t^k(M-m)^k}{k!} = \sum_{k = 1}^\infty \sum_{j \neq i} \ind_{ \sfd_\sfG(i,j) \le k }  \frac{t^k(M-m)^k}{k!} \le \sum_{k = 1}^\infty S k^\nu\,  \frac{t^k(M-m)^k}{k!}.
\end{equation}
Putting the above estimate back, we obtain (note $  \sum_{ k = 0 }^\infty \frac{t^k(M-m)^k}{k!} = \mee^{(M-m) t}$)
\begin{equation}
\begin{split}
    \sum_{j\in \sfV} \normo{ \nabla_{\bx_j} \bX_{t,i}^{\bx^{(j)}} } \le \mee^{-mt} + \mee^{-Mt} \sum_{k = 1}^\infty S k^\nu\,  \frac{t^k(M-m)^k}{k!},
\end{split}
\end{equation}
From \eqref{eqn:GradCtrl}, we obtain
\begin{equation}
\begin{split}
    \sum_{j\in \sfV} \normo{\nabla_j u(\bx^{(j)})} \le~& \sum_{j\in \sfV} \int_0^\infty \E \Rectbrac{ \normo{\nabla_{\bx_j} \bX_{t,i}^{\bx^{(j)}}} } \mdd t = \E \Rectbrac{ \int_0^\infty \sum_{j\in \sfV} \normo{\nabla_{\bx_j} \bX_{t,i}^{\bx^{(j)}}} \mdd t } \\
    \le~& \int_0^\infty \Brac{ \mee^{-mt} + S \mee^{-Mt} \sum_{k=1}^\infty k^\nu  \frac{t^k(M-m)^k}{k!} } \mdd t = \frac{1}{m} + \frac{S}{M} \sum_{k=1}^\infty k^\nu \Brac{ 1 - \frac{m}{M} }^k. 
\end{split}
\end{equation}
Recall that $\kappa = \frac{M}{m}$. Applying \Cref{lem:Li_ctrl}, it holds that
\begin{equation}
    \sum_{j\in \sfV} \normo{ \nabla_j u(\bx^{(j)})} \le \frac{1}{m} + \frac{S}{M} \nu! \Brac{\frac{m}{M}}^{-\nu-1} \Brac{ 1 - \frac{m}{M} } = \frac{1}{m} \Brac{ 1 + S \nu! \kappa^\nu  (1-\kappa^{-1})  } \le \frac{S \nu! \kappa^\nu}{m}.
\end{equation}
Taking supremum over $\bx^{(j)}$, we obtain the locality constant $\delta = \frac{S \nu! \kappa^\nu}{m}$.
\end{proof}

\subsection{Proof of Theorem~2}
\label{app:thm_diag_loc}

\begin{proof}[Proof of Theorem~2]

\Cref{lem:ExpSolu} holds because $\pi$ is $c$-strongly log-concave. As in the proof of \Cref{thm:MRF_loc}, the key is to control $\nabla_{\bx_j} \bX_{t,i}^\bx \in \mR^{d_i\times d_j}$ for $i\in[K]$ so that the gradient estimate in \eqref{eqn:GradCtrl} can be applied.

To estimate the $2$-norm of $\nabla_{\bx_j} \bX_{t,i}^\bx$, we introduce a unit test vector $\bv_j \in \mR^{d_j}$ such that $\norm{\bv_j} = 1$, and denote $\bg_t^\bx(i,j) = (\nabla_{\bx_j} \bX_{t,i}^\bx)  \bv_j \in \mR^{d_i}$ the matrix-vector product with this test vector. The function $\bg_t^\bx(i,j)$ then follows the differential equation (recall \eqref{eqn:pf_dXEvol}):
\begin{equation}
    \diff{}{t} \bg_t^\bx(i,j) = \diff{}{t} (\nabla_{\bx_j} \bX_{t,i}^\bx) \bv_j = - \sum_{\ell \in [K]} \sfH_{t}(i,\ell) \Brac{ \nabla_{\bx_j} \bX_{t,\ell}^\bx } \bv_j = - \sum_{\ell \in [K]} \sfH_t(i,\ell) \bg_t^\bx(\ell,j).
\end{equation}
Then, by the block-diagonal dominance assumption, we have
\begin{equation}
\begin{split}
    \frac{1}{2} \diff{}{t} \norm{ \bg_t^\bx(i,j) }^2 =~& - \bg_t^\bx(i,j)\matT \sum_{\ell \in [K]} \sfH_t(i,\ell) \bg_t^\bx(\ell,j) \\
    \le~& - A_{ii} \norm{ \bg_t^\bx(i,j) }^2 + \sum_{\ell\neq i} A_{i\ell} \norm{ \bg_t^\bx(i,j) } \norm{ \bg_t^\bx(\ell,j) }.
\end{split}
\end{equation}
Since the left-hand side of the above inequality is
\(
    \frac{1}{2} \diff{}{t} \norm{ \bg_t^\bx(i,j) }^2 =  \norm{ \bg_t^\bx(i,j) } \diff{}{t} \norm{ \bg_t^\bx(i,j) },
\)
we have the inequality
\begin{equation}
\label{eqn:pf_diag_ineq}
    \diff{}{t} \norm{ \bg_t^\bx(i,j) } \le - A_{ii} \norm{ \bg_t^\bx(i,j) } + \sum_{\ell \neq i} A_{i\ell} \norm{ \bg_t^\bx(\ell,j) }.
\end{equation}
The above inequality holds for any pair of indices $i,j \in [K]$, any test vector $\bv_j$, and any initial condition $\bx$ of the Langevin dynamics.
At $t=0$, we have
\begin{equation}
\label{eqn:pf_diag_init}
    \normo{ \bg_0^{\bx}(i,j) } =  \normo{ (\nabla_{\bx_j} \bx_i) \bv_j } = \ind_{i=j} \norm{\bv_j} = \ind_{i=j},
\end{equation}
for any $i,j \in [K]$ and any initial condition $\bx$. 
Since we seek a bound on $\sum_j \sup_{\bx} \normo{ \bg_t^\bx(i,j) }$, for each \(j\in[K]\), we choose a different initial condition $\bx^{(j)}$, and introduce the matrix $B_t \in \mR^{K\times K}$ defined by
\begin{equation}
    B_t(i,j) = \normo{ \bg_t^{\bx^{(j)}}(i,j) }, \quad \text{with} \quad B_0 = I_K.
\end{equation}
Then, the inequality \eqref{eqn:pf_diag_ineq} can be written compactly in matrix form as
\begin{equation}
    \diff{}{t} B_t \le - \tilde{A} B_t,
\end{equation}
where $\tilde{A}_{ij} := 2 A_{ii} \ind_{i=j} - A_{ij}$ as in \Cref{rem:l_cond} and the inequality $\le$ is defined entrywise.
By the $c$-diagonal dominance assumption on $A$, the matrix $\tilde{A} \in \mR^{K\times K}$ is $c$-diagonally dominant with non-positive off-diagonal entries (see \Cref{rem:l_cond}). Therefore, we can apply \Cref{lem:InfCtrl} to obtain
\begin{equation}
    \normo{B_t}_\infty \le \mee^{-c t} \normo  {B_0}_\infty = \mee^{-c t},
\end{equation}
which implies 
\(
    \max_{i\in[K]} \sum_{j \in [K]} \normo{ \nabla_{\bx_j} \bX_{t,i}^{\bx^{(j)}} \bv_j } \le \mee^{-c t}
\)
for arbitrary unit test vector $\bv_j$. This way, we obtain that
\begin{equation}
\label{eqn:pf_dXGrad2}
    \max_{i\in[K]} \sum_{j \in [K]} \normo{ \nabla_{\bx_j} \bX_{t,i}^{\bx^{(j)}} } \le \mee^{-c t}.
\end{equation}
Recall the gradient estimate \eqref{eqn:GradCtrl}, this implies 
\begin{equation}
\begin{split}
    \sum_{j \in [K]} \normo{ \nabla_j u(\bx^{(j)}) } \le~& \sum_{j \in [K]} \int_0^\infty \E \Rectbrac{ \normo{ \nabla_{\bx_j} \bX_{t,i}^{\bx^{(j)}} } } \mdd t \\
    =~& \E \Rectbrac{ \int_0^\infty \sum_{j \in [K]}  \normo{ \nabla_{\bx_j} \bX_{t,i}^{\bx^{(j)}} } \mdd t } \le \E  \Rectbrac{ \int_0^\infty \mee^{-c t} \mdd t } = c^{-1}.
\end{split}
\end{equation}
Taking supremum over $\bx^{(j)}$, we obtain the locality constant $\delta = c^{-1}$. 
\end{proof}

\subsection{Proof of Theorem~3}
\label{app:thm_Semigroup_loc}

\begin{proof}[Proof of Theorem~3]

(1) Let $f$ be such that
\(
    \norm{f}_{\infty,1} = \sum_{j\in[K]} \norm{ \nabla_j f }_{L^\infty} \le 1.
\)
Taking derivative with respect to $\bx_j$ on the definition \eqref{eqn:Pt}, we have 
\begin{equation}
    \nabla_j \sfP_t f(\bx) = \sum_{k \in [K]} \E [ \nabla_{\bx_j} \bX_{t,k}^\bx \cdot \nabla_k f(\bX_t^\bx) ].
\end{equation}
By the same argument as in the proof of \Cref{thm:MRF_loc}, we have (cf.~\eqref{eqn:pf_dXGrad})
\begin{equation}
    \normo{ \nabla_{\bx_j} \bX_{t,k}^\bx } \le \mee^{-M t} \sum_{q=\sfd_\sfG(j,k)}^\infty \frac{t^q(M-m)^q}{q!}. 
\end{equation}
Therefore, 
\begin{equation}
\begin{split}
    \norm{\nabla \sfP_t f}_{\infty,1} =~& \sum_{j\in[K]} \norm{ \nabla_j \sfP_t f }_{L^\infty} \le \sum_{j,k\in[K]} \E \Rectbrac{ \normo{ \nabla_{\bx_j} \bX_{t,k}^\bx } \norm{ \nabla_k f(\bX_t^\bx) } } \\
    \le~& \sum_{k\in[K]} \norm{ \nabla_k f }_{L^\infty} \Rectbrac{ \sum_{j\in[K]} \mee^{-M t} \sum_{q=\sfd_\sfG(j,k)}^\infty \frac{t^q(M-m)^q}{q!} }.
\end{split}
\end{equation}
Using the $(S,\nu)$-locality, we have for any fixed $k\in[K]$,
\begin{equation}
\begin{split}
    \sum_{j\in[K]} \sum_{q=\sfd_\sfG(j,k)}^\infty \frac{t^q(M-m)^q}{q!} =~& \sum_{q=0}^\infty \frac{t^q(M-m)^q}{q!} + \sum_{q=1}^\infty \sum_{j:1\le \sfd_\sfG(j,k) \le q} \frac{t^q(M-m)^q}{q!} \\
    \le~& \mee^{(M-m)t} + \sum_{q=1}^\infty S q^\nu \frac{t^q(M-m)^q}{q!} = \mee^{(M-m)t} \Rectbrac{ 1 + S p_\nu ( t(M-m) ) }, 
\end{split}
\end{equation}
where we denote 
\(
    p_\nu(x) := \mee^{-x} \sum_{q=0}^\infty q^\nu \frac{x^q}{q!},
\)
which is a monic polynomial of degree $\nu$ by \Cref{lem:pk}. Putting the above estimate back, we obtain
\begin{equation}
    \norm{\nabla \sfP_t f}_{\infty,1} \le \sum_{k\in[K]} \norm{ \nabla_k f }_{L^\infty} \mee^{-mt} \Rectbrac{ 1+ S p_\nu ( t(M-m) ) } = \mee^{-mt} \Rectbrac{ 1+ S p_\nu ( t(M-m) ) } \norm{\nabla f}_{\infty,1}.
\end{equation}
(2) The proof is similar to (1). It suffices to replace \eqref{eqn:pf_dXGrad} with \eqref{eqn:pf_dXGrad2}, i.e.,
\begin{equation}
    \max_{k\in[K]} \sum_{j \in [K]} \normo{ \nabla_{\bx_j} \bX_{t,k}^\bx } \le \mee^{-c t}. 
\end{equation}
Then similarly we have
\begin{equation}
    \norm{ \nabla \sfP_t f}_{\infty,1} \le \sum_{j,k\in[K]} \E \Rectbrac{ \norm{ \nabla_{\bx_j} \bX_{t,k}^\bx } \norm{ \nabla_k f(\bX_t^\bx) } } \le \mee^{-ct} \sum_{k\in[K]} \norm{ \nabla_k f }_{L^\infty} = \mee^{-ct} \norm{\nabla f}_{\infty,1}.
\end{equation}
This completes the proof. 
\end{proof}

\subsection{Proof of Theorem~4}
\label{app:thm_margin}

\begin{proof}[Proof of Theorem~4]
By Kantorovich duality \cite{MR2459454}, the $1$-Wasserstein distance between marginal densities can be expressed as
\begin{equation}
    \sfW_1(\pi_i^{},\pi_i') = \sup_{\phi \in \Lip_1(\mR^{d_i})} \int \phi (\bx_i) \Brac{ \pi_i(\bx_i) - \pi_i' (\bx_i)} \mdd \bx_i, \quad \forall i \in [K],
\end{equation}
where $\Lip_1(\mR^{d_i})$ denotes the class of $1$-Lipschitz test functions $\phi: \mR^{d_i} \rightarrow \mR$ defined for the block of variables $\bx_i$.
For a given block index $i \in [K]$ and any corresponding test function $\phi \in \Lip_1(\mR^{d_i})$, it holds that
\begin{equation}
\begin{split}
    \int \phi (\bx_i) \Brac{ \pi_i^{} (\bx_i) - \pi_i' (\bx_i)} \mdd \bx_i =~& \int \Brac{ \phi (\bx_i) - \E_{\pi_i'}[\phi(\bx_i)] } \pi_i(\bx_i) \mdd \bx_i \\
    =~& \int \Brac{ \phi (\bx_i) - \E_{\pi'}[\phi(\bx_i)] } \pi(\bx) \mdd \bx,
\end{split}
\end{equation}
where $\E_{\pi_i'}[\phi(\bx_i)] = \E_{\pi'}[\phi(\bx_i)]$ is a constant.
Let $u(\bx)$ solve the marginal Stein equation
\begin{equation}
    \Delta u(\bx) + \nabla \log \pi' (\bx) \cdot \nabla u(\bx) = \phi(\bx_i) - \E_{\pi'} [\phi(\bx_i)].
\end{equation}
Applying integration by parts, for any $v \in H^1(\pi')$, we have
\begin{equation}
\label{eqn:pf_marginal_intbyparts}
    \int v(\bx) \Brac{ \phi(\bx_i) - \E_{\pi'} [\phi(\bx_i)] } \pi' (\bx) \mdd \bx = - \int \nabla v(\bx) \cdot \nabla u(\bx) \pi' (\bx) \mdd \bx.
\end{equation}
Here we do not require $v$ to be mean-zero since adding a constant to $v$ does not change the above equality.
By the $\delta$-locality assumption, the solution $u(\bx)$ satisfies the gradient estimate
\begin{equation}
\label{eqn:pf_marginal_grad_est}
    \sum_{j=1}^K \norm{\nabla_j u}_{L^\infty} \le \delta.
\end{equation}
Since $C_c^1(\mR^d) \subseteq H^1(\pi')$, \eqref{eqn:pf_marginal_intbyparts} holds for all $v \in C_c^1(\mR^d)$. By passing to the limit, it holds for all $v \in W^{1,1}(\pi')$. In particular, we can take $v(\bx) = \frac{\pi(\bx)}{\pi'(\bx)}$ if $v(\bx) \in W^{1,1}(\pi')$, and then
\begin{equation}
\begin{split}
    \int \Brac{ \phi (\bx_i) - \E_{\pi'}[\phi(\bx_i)] } \pi(\bx) \mdd \bx =~& - \int \nabla \Brac{ \frac{\pi(\bx)}{\pi'(\bx)} } \cdot \nabla u(\bx) \pi' (\bx) \mdd \bx \\
    =~& - \int \Brac{ \nabla \log \pi(\bx)- \nabla \log \pi'(\bx) } \cdot \nabla u(\bx) \pi (\bx) \mdd \bx  \\
    =~& \sum_{j=1}^K \int \nabla_j u(\bx) \cdot \Brac{ \nabla_j \log \pi' (\bx) - \nabla_j \log \pi (\bx) } \pi(\bx) \mdd \bx \\
    \le~& \sum_{j=1}^K \norm{\nabla_j u}_{L^\infty} \int \norm{ \nabla_j \log \pi' (\bx) - \nabla_j \log \pi (\bx) } \pi(\bx) \mdd \bx \\
    \le~& \delta \; \max_{j \in [K]} \int \norm{ \nabla_j \log \pi' (\bx) - \nabla_j \log \pi (\bx) } \pi(\bx) \mdd \bx.
\end{split}
\end{equation}
Here we use the gradient estimate \eqref{eqn:pf_marginal_grad_est}, and note the final bound is finite if and only if $\frac{\pi(\bx)}{\pi'(\bx)} \in W^{1,1}(\pi')$.
If $\frac{\pi(\bx)}{\pi'(\bx)} \not\in W^{1,1}(\pi')$, the right-hand side is infinite, so the inequality holds trivially.
The above inequality holds for every $i \in [K]$, and thus we obtain 
\begin{equation}
\begin{split}
    \max_{i\in[K]} \sfW_1(\pi_i^{},\pi_i') \le~& \max_{i\in[K]} \sup_{\phi\in\Lip_1(\mR^{d_i})} \int \phi(\bx_i) \Brac{ \pi_i^{}(\bx_i) - \pi_i'(\bx_i) } \mdd \bx_i \\
    \le~& \max_{i\in[K]} \sup_{\phi \in\Lip_1(\mR^{d_i})} \Brac{ \delta \; \max_{j \in [K]} \int \norm{ \nabla_j \log \pi' (\bx) - \nabla_j \log \pi (\bx) } \pi(\bx) \mdd \bx } \\
    =~& \delta \; \max_{j \in [K]} \norm{ \nabla_j \log \pi' - \nabla_j \log \pi }_{L^1(\pi)}.
\end{split}
\end{equation}
This completes the proof.
\end{proof}

\subsection{Proof of Theorem~5}
\label{app:thm_margin_multi}

\begin{proof}[Proof of Theorem~5]
For any index set $\mcJ\subset [K]$, we have the $1$-Wasserstein distance between the corresponding marginals
\begin{equation}
    \sfW_1(\pi_\mcJ^{},\pi_\mcJ') = \sup_{\phi \in \Lip_1(\mR^{d_\mcJ})} \int \phi(\bx_\mcJ) \Brac{ \pi_\mcJ^{}(\bx_\mcJ) - \pi_\mcJ' (\bx_\mcJ) } \mdd \bx_\mcJ.
\end{equation}
For a given $1$-Lipschitz test function $\phi_\mcJ : \mR^{d_\mcJ} \rightarrow \mR$, let $u(\bx)$ solve the marginal Stein equation
\begin{equation}
    \Delta u(\bx) + \nabla \log \pi' (\bx) \cdot \nabla u(\bx) = \phi(\bx_\mcJ) - \E_{\pi'} [\phi(\bx_\mcJ)].  
\end{equation}
Using the same argument as in the proof of \Cref{lem:ExpSolu}, it holds that
\begin{equation} 
    \nabla_j u(\bx) = - \int_0^\infty \E \Rectbrac{ (\nabla_{\bx_j} \bX_{t,\mcJ}^\bx)^\top \nabla \phi (\bX_{t,\mcJ}^\bx) } \mdd t,
\end{equation}
where $\bX_t^\bx$ is the path solution of the Langevin dynamics \eqref{eqn:Langevin} for $\pi'$ with initial condition $\bx$, and thus
\begin{equation}
    \norm{ \nabla_j u(\bx) } \le \int_0^\infty \E \Rectbrac{ \normo{ \nabla_j \bX_{t,\mcJ}^\bx } \norm{ \nabla \phi (\bX_{t,\mcJ}^\bx) }  } \mdd t \le \int_0^\infty \E \Rectbrac{ \normo{ \nabla_j \bX_{t,\mcJ}^\bx } } \mdd t.
\end{equation}
Since $\normo{ \nabla_j \bX_{t,\mcJ}^\bx } \le \sum_{i\in \mcJ} \normo{ \nabla_j \bX_{t,i}^\bx }$, we obtain that 
\begin{equation}
    \norm{\nabla u}_{\infty,1} = \sum_{j\in[K]} \norm{ \nabla_j u(\bx) } \le \sum_{i\in\mcJ} \sum_{j\in [K]} \E \Rectbrac{ \int_0^\infty \normo{ \nabla_j \bX_{t,i}^\bx } \mdd t }.
\end{equation}
Depending on which class $\pi'$ belongs to, and following the proofs of either \Cref{thm:MRF_loc} or \Cref{thm:diag_loc}, we obtain the bound
\begin{equation}
    \max_{i\in \mcJ} \sum_{j\in [K]} \E \Rectbrac{ \int_0^\infty \normo{ \nabla_j \bX_{t,i}^\bx } \mdd t } \le \delta.
\end{equation}
Therefore, we have the inequality
\(
\norm{\nabla u}_{\infty,1} \le \delta |\mcJ|.
\)
Following the same argument as in the proof of \Cref{thm:margin} and applying the above inequality, we obtain 
\begin{equation}
\begin{split}
    \sfW_1(\pi_\mcJ^{},\pi_\mcJ') \le~& \norm{\nabla u}_{\infty,1} \cdot \max_{j \in [K]} \norm{ \nabla_j \log \pi' - \nabla_j \log \pi }_{L^1(\pi)} \\
    \le~& \delta |\mcJ|  \max_{j \in [K]} \norm{ \nabla_j \log \pi' - \nabla_j \log \pi }_{L^1(\pi)}.
\end{split}
\end{equation}
This completes the proof. 
\end{proof}

\section{Proofs of the auxiliary lemmas}
\label{app:lemmas_1}

\subsection{Proof of Lemma~2}
\label{app:diff_ctrl}

\begin{proof}
The proof is divided into four steps.
\begin{itemize}
    \item First, we argue by scaling that it suffices to prove the result for $t=1$.
    \item Next, we show that the solution to \eqref{eqn:MatODE} can be represented by a Dyson series.
    \item Then, we represent the Dyson series using an equivalent form.
    \item Finally, we derive the desired bound by using the equivalent representation.
\end{itemize}
We comment that the key idea is to use the banded matrix approximation trick in \cite{MR758197}. The technical difficulty here is to generalize to time-dependent matrices. Since these matrices are no longer commutative, we need to use the Dyson series representation.
\ \\[5pt]
\noindent
{\bf I. Scaling.}
Assume that we have proved the result for $t=1$. Now for general $t_0 > 0$, let $\sfX_t$ solve \eqref{eqn:MatODE}, then $\sfX_{t_0 s}$ solves
\[
    \diff{}{s} \sfX_{t_0s} = - t_0 \sfH_{t_0s} \sfX_{t_0s}, \quad \sfX_0 = I_d.
\]
For the rescaled system, we obtain that at $s=1$ (note $\norm{t_0\sfH_{t_0s}} \le Mt_0$),
\[
    \normo{ \sfX_{t_0}(i,j) } \le \exp(-Mt_0) \sum_{k=\sfd_\sfG(i,j)}^\infty \frac{M^k t_0^k}{k!}.
\]
This proves the result for general $t_0$. Hence, in the following, we only consider the case $t=1$.
\ \\[5pt]
{\bf II. Dyson series solution.}
By variation of constants, we have
\[
    \sfX_t = I_d - \int_0^t \sfH_s \sfX_s \mdd s.
\]
Applying this identity recursively, we obtain
\begin{align*}
    \sfX_t =~& I_d - \int_0^t \sfH_s \Brac{ I_d - \int_0^s \sfH_u \sfX_u \mdd u } \mdd s \\
    =~& I_d - \int_0^t \sfH_s \mdd s + \int_0^t \int_0^s \sfH_s \sfH_u \sfX_u \mdd u \mdd s \\ %= \cdots \\
    =~& I_d + \sum_{n=1}^{N-1} (-1)^n \int_{[0,t]^n} \sfH_{t_n} \cdots \sfH_{t_1} \ind_{t_1\le t_2 \le \cdots \le t_n} \mdd t_1 \cdots \mdd t_n \\
    & + (-1)^N \int_{[0,t]^N} \sfH_{t_N} \cdots \sfH_{t_1} \sfX_{t_1} \ind_{t_1\le t_2 \le \cdots \le t_N} \mdd t_1 \cdots \mdd t_N.
\end{align*}
For simplicity, denote 
\[
    X_0(t) = I_d, \quad X_n(t)  := \frac{n!}{t^n} \int_{[0,t]^n} \sfH_{t_n} \cdots \sfH_{t_1} \ind_{t_1\le t_2 \le \cdots \le t_n} \mdd t_1 \cdots \mdd t_n, \quad n \ge 1.
\]
\[
    R_N(t) = \int_{[0,t]^N} \sfH_{t_N} \cdots \sfH_{t_1} \sfX_{t_1} \ind_{t_1\le t_2 \le \cdots \le t_N} \mdd t_1 \cdots \mdd t_N.
\]
Then 
\[
    \sfX_t = \sum_{n=0}^{N-1} (-1)^n \frac{t^n}{n!} X_n(t)+(-1)^N R_N(t).
\]
Now we prove that $\lim_{N\gtoinf} R_N(t) = 0$. First note that $\norm{\sfX_t}_2 \le 1$, since
\[
    \diff{}{t} \sfX_t\matT \sfX_t = - 2 \sfX_t\matT \sfH_t \sfX_t ~\St~ \sfX_t\matT \sfX_t = I_d - \int_0^t \sfX_s\matT \sfH_s \sfX_s \mdd s \preceq I_d.
\]
Therefore, as $N\gtoinf$,
\begin{align*}
    \norm{R_N(t)}_2 \le~& \int_{[0,t]^N} \norm{\sfH_{t_N}}_2 \cdots \norm{\sfH_{t_1}}_2 \norm{\sfX_{t_1}}_2 \ind_{t_1\le t_2 \le \cdots \le t_N} \mdd t_1 \cdots \mdd t_N \\
    \le~& M^N \int_{[0,t]^N} \ind_{t_1\le t_2 \le \cdots \le t_N} \mdd t_1 \cdots \mdd t_N = \frac{M^N t^N}{N!} \to 0. 
\end{align*}
This proves that the Dyson series converges, and
\[
    \sfX_t = \sum_{n=0}^\infty (-1)^n \frac{t^n}{n!} X_n(t).
\]
\ \\
{\bf III. Representation of polynomials.}
Denote the matrix process space 
\[
    \msX = \text{span}\{X_n(t), n\ge 0\} = \mean{ \Big\{ \sum_{k=0}^n a_k X_k(t) : \forall k \in [n],~ a_k\in\mC;~ n \in \mN \Big\} }.
\]
We define the representation of any polynomial $P$ in $\msX$ as 
\[
    P[X](t) := \sum_{k=0}^n a_k X_k(t), \quad \text{if } P(x) = \sum_{k=0}^n a_k x^k.
\]
We will show later that the representation has an equivalent definition:
\begin{equation}
\label{eqn:poly_rep}
    P\{X\}(t) := \frac{a_n}{t^n} \sum_{\sigma \in S_n} \int_{[0,1]^n} (\sfH_{t_n} - x_{\sigma_n} I_d) \cdots (\sfH_{t_1} - x_{\sigma_1} I_d) \ind_{t_1\le t_2 \le \cdots \le t_n} \mdd t_n \cdots \mdd t_2 \mdd t_1,
\end{equation}
\[
    \text{if } P(x) = a_n \prod_{k=1}^n (x-x_k).
\]
Here $S_n$ is the set of all permutations of $[n]$.
We denote $P\{X\}$ to distinguish the two definitions. Note that the representation can be extended from polynomials to any analytic functions.
We next verify the equivalence by induction on $n$. First note $n=1$ is obvious, 
\[
    (a_1(x-x_1)) \{X\} (t) = \frac{a_1}{t} \int_0^t (\sfH_{t_1} - x_1 I_d) \mdd t_1 = a_1 ( X_1(t) - x_1 I_d ) = (a_1(x-x_1)) [X] (t).
\]
Now consider $n\ge 2$. Note that $P[X](t)$ and $P\{X\}(t)$ can both be viewed as multilinear maps on $x_1,\dots,x_n$. Computing the partial derivative with respect to $x_n$, we obtain
{\small
\begin{align*}
    & \nabla_{x_n} \Brac{ P\{X\}(t) } \\
    =~& \frac{a_n}{t^n} \sum_{k=1}^n \sum_{\sigma\in S_n, \sigma_k = n} \int_{[0,1]^n} (\sfH_{t_n} - x_{\sigma_n} I_d) \cdots \nabla_{x_n} (\sfH_{t_k} - x_n I_d) \cdots (\sfH_{t_1} - x_{\sigma_1} I_d) \ind_{t_1\le t_2 \le \cdots \le t_n} \mdd t_n \cdots \mdd t_1 \\
    =~& \frac{a_n}{t^n} \sum_{k=1}^n \sum_{\sigma\in S_n, \sigma_k = n} \int_{[0,1]^n} (\sfH_{t_n} - x_{\sigma_n} I_d) \cdots (- I_d) \cdots (\sfH_{t_1} - x_{\sigma_1} I_d) \ind_{t_1\le t_2 \le \cdots \le t_n} \mdd t_n \cdots \mdd t_1 \\
    =~& \frac{a_n}{t^n} \sum_{k=1}^n \sum_{\sigma\in S_n, \sigma_k = n} \int_{[0,1]^{n-1}} (\sfH_{t_n} - x_{\sigma_n} I_d) \cdots ( t_{k-1} - t_{k+1} ) \cdots (\sfH_{t_1} - x_{\sigma_1} I_d) \ind_{t_1\le t_2 \le \cdots \le t_n} \mdd t_n \cdots \mdd \hat{t}_k \cdots \mdd t_1 \\
    =~& \frac{a_n}{t^n} \sum_{k=1}^n \sum_{\sigma \in S_{n-1}} \int_{[0,1]^{n-1}} (\sfH_{t_{n-1}} - x_{\sigma_{n-1}} I_d) \cdots ( t_{k-1} - t_k ) \cdots (\sfH_{t_1} - x_{\sigma_1} I_d) \ind_{t_1\le t_2 \le \cdots \le t_{n-1}} \mdd t_{n-1} \cdots \mdd t_1 \\
    =~& \frac{a_n}{t^n} \cdot (- t) \sum_{\sigma \in S_{n-1}} \int_{[0,1]^{n-1}} (\sfH_{t_{n-1}} - x_{\sigma_{n-1}} I_d) \cdots (\sfH_{t_1} - x_{\sigma_1} I_d) \ind_{t_1\le t_2 \le \cdots \le t_{n-1}} \mdd t_{n-1} \cdots \mdd t_1 \\
    =~& - a_n \Brac{ \prod_{k=1}^{n-1} (x-x_k) } \{X\}(t) = - a_n \Brac{ \prod_{k=1}^{n-1} (x-x_k) } [X](t).
\end{align*}}

\noindent
Here the first equality follows from the position of $x_n$. The third equality follows by integrating against the variable $t_k$ and using the constraint $t_{k-1} \le t_k \le t_{k+1}$. Here we set $t_{n+1}=1$. The fourth equality follows by relabeling the index $(k+1,\dots,n+1) \mapsto (k,\dots,n)$. The last equality follows from the induction hypothesis. By symmetry, the relation holds for any $x_i$:
\[
    \nabla_{x_i} \Brac{ P\{X\}(t) } = - \Big( a_n \prod_{j\neq i} (x-x_j) \Big) [X](t) = (\nabla_{x_i} P) [X] (t), 
\]
since $\nabla_{x_i} P(x) = - a_n \prod_{j \neq i} (x-x_j)$. Since the representation $P[X]$ is linear in the coefficients of $P$, it is direct to verify that
\[
    \nabla_{x_i} ( P[X] (t) ) = (\nabla_{x_i} P) [X] (t) = \nabla_{x_i} \Brac{ P\{X\}(t) }.
\]
Finally, note that when $x_1 = \cdots = x_n = 0$,
\[
    x^n\{X\}(t) = \frac{n!}{t^n} \int_{[0,1]^n} \sfH_{t_n} \cdots \sfH_{t_1} \ind_{t_1\le t_2 \le \cdots \le t_n} \mdd t_1 \cdots \mdd t_n = X_n(t) = x^n[X](t).
\]
The two multilinear maps agree at one point, as do all their partial derivatives. Therefore, they must be identical. This shows that the equivalence holds for $n$, completing the induction.
\ \\[5pt]
{\bf IV. Banded matrix approximation.}
By Taylor expansion, 
\[
    \exp(-x) = \exp(-M) \sum_{n=0}^\infty \frac{ (M-x)^n }{n!}.
\]
Representing the series in $\msX$, we obtain (denote $\hat{X}_n(x) = (M-x)^n$)
\[
    \sfX_1 = \exp(-M) \sum_{n=0}^\infty  \frac{1}{n!} \hat{X}_n [X] (1),
\]
\[
    \hat{X}_n[X](1) = \sum_{\sigma \in S_n} \int_{[0,1]^n} (MI_d-\sfH_{t_n}) \cdots (MI_d-\sfH_{t_1}) \ind_{t_1\le t_2 \le \cdots \le t_n} \mdd t_1 \cdots \mdd t_n.
\]
Here we use the alternative representation \eqref{eqn:poly_rep}. We can truncate the Dyson series of $\sfX_1$ as
\[
    \sfX_1 = \exp(-M) \sum_{q=0}^{n}  \frac{1}{q!} \hat{X}_q [X] (1) + \exp(-M) \sum_{q=n+1}^\infty \frac{1}{q!} \hat{X}_q [X] (1).
\]
Consider the off-diagonal entry $\sfX_1(i,j)$. Take $n = \sfd_\sfG(i,j) - 1 $, then since all the path in $\sfG$ connecting $i$ and $j$ has length no less than $\sfd_\sfG(i,j)>n$, it must hold that
\[
    \forall 1\le q \le n, \quad \Rectbrac{  (MI_d-\sfH_{t_q}) \cdots (MI_d-\sfH_{t_1}) } (i,j) = 0 ~\St~ \hat{X}_q[X](1)(i,j) = 0,
\]
Therefore, 
\[
    \sfX_1(i,j) = \exp(-M) \sum_{q=n+1}^\infty \frac{1}{q!} \hat{X}_q [X] (1)(i,j).
\]
Since $0 \preceq \sfH_{t_k} \preceq M I$, it holds that $\norm{MI - \sfH_{t_k}}_{\rm op} \le M$, and thus
\[
    \normo{ \hat{X}_n[X](1) }_{\rm op} \le M^n n! \int_{[0,1]^n} \ind_{t_1\le t_2 \le \cdots \le t_n} \mdd t_1 \cdots \mdd t_n = M^n. 
\]
\[
    \St~ \normo{\sfX_1(i,j)} \le \exp(-M) \sum_{q=n+1}^\infty \frac{M^q}{q!} = \exp(-M) \sum_{q=\sfd_\sfG(i,j)}^\infty \frac{M^q}{q!}.
\]
This verifies the case $i\neq j$. For $i=j$, the result follows from $\normo{\sfX_1(i,i)} \le \norm{\sfX_1}\le 1$.
\end{proof}

\subsection{Auxiliary summation bound}
\label{app:Li_ctrl}

\begin{lem}
\label{lem:Li_ctrl}
For any $t\ge 0$ and $x\in(0,1)$, it holds that
\(
    \sum_{k\ge 1} k^t (1-x)^k < 2 \Gamma(t+1) x^{-t-1} (1-x),
\)
where $\Gamma(\cdot)$ is the Gamma function. For integer-valued $t\in\mN$, the bound can be improved to
\(
    \sum_{k\ge 1} k^t (1-x)^k \le t! x^{-t-1} (1-x).
\)
\end{lem}

\begin{proof}[Proof of \Cref{lem:Li_ctrl}]
We prove by induction. For $t=0$, it holds that 
\[
    \sum_{k\ge 1} (1-x)^k = \frac{1-x}{x}. 
\]
For $t\in(0,1)$, applying the Abel transformation, we have
\[
    \sum_{k\ge 1} k^t (1-x)^k = x^{-1} \sum_{k\ge 1} k^t \Rectbrac{ (1-x)^k - (1-x)^{k+1} } = x^{-1} \sum_{k\ge 1} \Brac{ k^t - (k-1)^t } (1-x)^k. 
\]
Since $k^t - (k-1)^t \le t (k-1)^{t-1}$ when $t\in(0,1)$ and $k\ge 2$, we have
\begin{align*}
    \sum_{k\ge 1} k^t (1-x)^k \le~& x^{-1} \Rectbrac{ (1-x) + \sum_{k\ge 2} t (k-1)^{t-1} (1-x)^k } \\
    \le~& x^{-1} (1-x) \Rectbrac{ 1 + t \sum_{k\ge 2} (k-1)^{t-1} \mee^{-(k-1)x} } \\
    \le~& x^{-1} (1-x) \Rectbrac{ 1 + t \int_0^\infty y^{t-1} \mee^{-yx} \mdd y } \\
    =~& x^{-1} (1-x) \Brac{ 1 + t \Gamma(t) x^{-t} } < 2 \Gamma(t+1) x^{-t-1} (1-x),
\end{align*}
where we use the identity $t\Gamma(t) = \Gamma(t+1)$ and $\Gamma(t+1)x^{-t}>1$ in the last step. This verifies the case $t\in[0,1)$. Suppose the inequality holds for $t-1\ge 0$, then using the Abel transformation again, we have
\begin{align*}
    \sum_{k\ge 1} k^t (1-x)^k =~& x^{-1} \sum_{k\ge 1} \Brac{ k^t - (k-1)^t } (1-x)^k \\
    \le ~& x^{-1} \sum_{k\ge 1} t k^{t-1} (1-x)^k \\
    <~& x^{-1} t \cdot 2 \Gamma(t) x^{-(t-1)-1} (1-x) \\ =~& 2 \Gamma(t+1) x^{-t-1} (1-x),
\end{align*}
where the first inequality follows by the elementary inequality $k^t - (k-1)^t \le t k^{t-1}$ when $t\ge 1$, and the second inequality follows by induction hypothesis. The simplified inequality for $t\in\mN$ can be obtained in the same way. This completes the proof. 
\end{proof}

\subsection{Auxiliary infinity-norm comparison}
\label{app:InfCtrl}

\begin{lem}
\label{lem:InfCtrl}
Suppose $B_t \in \mR_+^{K\times K}$ is a time-dependent entrywise nonnegative matrix satisfying
\begin{equation}
    \diff{}{t} B_t \le - \tilde{A} B_t,
\end{equation}
where the inequality $\le$ is defined entrywise, and $\tilde{A} \in \mR^{K\times K}$ is $c$-diagonally dominant with non-positive off-diagonal entries, i.e.,
\(
    \sum_{j\neq i} |\tilde{A}_{ij}| + c \le \tilde{A}_{ii};\, \tilde{A}_{ij} \le 0
\)
for all $ i,j \in [K]$ with $i\neq j$. Then for any $t\ge 0$, it holds that
\begin{equation}
    \normo{B_t}_\infty \le \mee^{-c t} \normo{B_0}_\infty.
\end{equation}
Here $\normo{B}_\infty := \sup_{\bv\neq 0} \frac{\norm{B\bv}_\infty }{ \norm{\bv}_\infty }$ denotes the matrix infinity norm.
\end{lem}

\begin{proof}[Proof of \Cref{lem:InfCtrl}]
Denote $B_t^c := \mee^{c t} B_t$, then we have
\begin{equation}\label{eqn:lem4_eq1}
    \diff{}{t} B_t^c = \mee^{c t} \Brac{ \diff{}{t} B_t + c B_t } \le \mee^{c t} \Brac{ - \tilde{A} B_t + c B_t } = \Brac{ - \tilde{A} + c I_K } B_t^c.
\end{equation}
Multiplying both sides of the above inequality by $\ones = (1,\dots,1)\matT \in \mR^K$ from right, we have
\[
    \diff{}{t} ( B_t^c \ones ) \le \Brac{ - \tilde{A}  + c I_K } B_t^c \ones,
\]
which preserves the inequality \eqref{eqn:lem4_eq1} since it essentially takes summation over columns.%\todo{not rows, should be columns} 
We then want to verify the following claim:
\begin{equation}    \label{eqn:pf_Mat1norm}
   \normo{ B_t^c \ones }_\infty \le  \normo{ B^c_0 \ones }_\infty, \quad  \forall t \ge 0.
\end{equation}
Since $B_t \in \mR_+^{K\times K}$ by construction, we have
\[
    \mee^{c t} \normo{B_t}_\infty = \normo{B_t^c}_\infty = \max_{i\in[K]} \sum_{j \in [K]} B_t^c(i,j) = \normo{ B_t^c \ones }_\infty,
\]
and thus the claim \eqref{eqn:pf_Mat1norm} is equivalent to the conclusion $\norm{B_t}_\infty \le \mee^{-c t} \norm{B_0}_\infty$.

We prove the claim \eqref{eqn:pf_Mat1norm} by contradiction. Suppose \eqref{eqn:pf_Mat1norm} is false, then there exists some $s\ge 0$ and $i$ such that
\[
    \diff{}{t} (B^c_{s} \ones)_i  > 0 \quad \text{and}\quad (B^c_{s} \ones)_i = \normo{ B^c_{s} \ones }_\infty. 
\]
Since $\tilde{A} \in \mR^{K\times K}$ is $c$-diagonally dominant with non-positive off-diagonal entries, we have
\begin{align*}
    \diff{}{t} (B^c_{s} \ones)_i \le~& \Brac{ - \tilde{A}_{ii} + c } (B^c_{s} \ones)_i + \sum_{j\neq i} (-\tilde{A}_{ij}) (B^c_{s} \ones)_j  \\
    \le~& \Brac{ - \tilde{A}_{ii} + c } (B^c_{s} \ones)_i + \sum_{j\neq i} (-\tilde{A}_{ij}) (B^c_{s} \ones)_i \\
    =~& \Brac{ - \tilde{A}_{ii} + c + \sum_{j\neq i} |\tilde{A}_{ij}| } (B^c_{s} \ones)_i \le 0,
\end{align*}
which leads to a contradiction. This proves our result. 
\end{proof}

\subsection{Polynomial identity}
\label{app:pk}

\begin{lem}
\label{lem:pk}
The function
$p_\nu(x) := \mee^{-x} \sum_{q=0}^\infty q^\nu \frac{x^q}{q!}$ is a monic polynomial of degree $\nu$.
\end{lem}

\begin{proof}[Proof of \Cref{lem:pk}]
By the definition of the function $p_\nu$, we have 
\[
    \mee^x p_\nu(x) = \sum_{q=0}^\infty q^\nu \frac{x^q}{q!},
\] 
Using the identity  
\[
    \sum_{q=0}^\infty q^\nu \frac{x^q}{q!} = \Brac{ x \diff{}{x} }^\nu \Brac{ \sum_{q=0}^\infty \frac{x^q}{q!} } = \Brac{ x \diff{}{x} }^\nu \mee^x,
\]
we have 
\[
\mee^x p_\nu(x) =\Brac{ x \diff{}{x} }^\nu \mee^x,
\]
so that
\[
    \mee^x p_{\nu+1}(x) = \Brac{ x \diff{}{x} } \Brac{ \mee^x p_\nu(x) } = x \mee^x \Brac{ p_\nu(x) + p_\nu'(x) }.
\]
This defines the recursion,
\[
    p_{\nu+1}(x) = x \Brac{ p_\nu(x) + p_\nu'(x) }, \quad p_0(x) = 1.
\]
Applying induction, we can conclude that the function $p_\nu$ is a monic polynomial of degree $\nu$.
\end{proof}

\section{Proof of Theorem~6}
\label{app:LLIS}

\subsection{Auxiliary lemmas for Theorem~6}
\label{app:LLIS_lemmas}

To prove \Cref{thm:LLIS_approx}, we need the following two lemmas.

\begin{lem}
\label{lem:MatInvLoc}
Suppose a symmetric matrix $\sfQ \in \mR^{d\times d}$ has dependency graph $\sfG$, i.e., $\sfQ(i,j) = 0$ if $i \notin \mcN(j)$.
Assume that $m I \preceq \sfQ \preceq M I$ for some $0<m\le M <\infty$. Then 
\[
    \norm{ \sfQ^{-1}(i,j) }_2 \le \frac{1}{m} \Brac{ 1 - \frac{m}{M} }^{\sfd_\sfG(i,j)}. 
\]
\end{lem}

\begin{proof}
Since $\sfQ\succeq m I $ for $m>0$, it holds that
\[
    \sfQ^{-1} = \int_0^\infty \mee^{-t \sfQ} \mdd t = \int_0^\infty \mee^{-mt} \mee^{-t (\sfQ-mI)} \mdd t.
\]
Note $\sfX_t := \mee^{-t(\sfQ-mI)}$ solves the ODE $ \diff{}{t} \sfX_t = - (\sfQ-mI) \sfX_t $.
By \Cref{lem:diff_ctrl}, it holds that 
\[
    \norm{ [\mee^{-t(\sfQ-mI)}] (i,j) }_2 \le \exp(-t(M-m)) \sum_{k=\sfd_\sfG(i,j)}^\infty \frac{t^k (M-m)^k}{k!}.
\]
Therefore, 
\begin{align*}
    \norm{ \sfQ^{-1}(i,j) }_2 \le~& \int_0^\infty \mee^{-mt} \norm{ [\mee^{-t(\sfQ-mI)}] (i,j) }_2 \mdd t \\
    \le~& \sum_{k=\sfd_\sfG(i,j)}^\infty \int_0^\infty \mee^{-Mt} \frac{t^k (M-m)^k}{k!} \mdd t \\
    =~& \frac{1}{M} \sum_{k=\sfd_\sfG(i,j)}^\infty \Brac{ 1 - \frac{m}{M} }^k = \frac{1}{m} \Brac{ 1 - \frac{m}{M} }^{\sfd_\sfG(i,j)}. 
\end{align*}
This completes the proof.
\end{proof}

\begin{lem}
\label{lem:Kl1Bound}
Under \Cref{asm:LLIS}, and denote $\sfK = \sfC_{r,r}^\dag \sfC_{r,\bot}$, where $\sfC_{\bot,r} = \Pi_{\bot} \sfC \Pi_{r}$ and $\sfC_{r,r} = \Pi_{r} \sfC \Pi_{r}$
Then it holds that 
\[
    \sum_{k\in[K]} \norm{ \sfK(j,k) } \le S(1+S) \nu! \kappa^{\nu+2}. 
\]
\end{lem}

\begin{proof}[Proof of \Cref{lem:Kl1Bound}]

Denote $\sfQ = \sfC^{-1}$, and similarly define 
\[
    \sfQ_{r,\bot} = \Pi_r \sfQ \Pi_\bot, \quad \sfQ_{\bot,\bot} = \Pi_\bot \sfQ \Pi_\bot.
\]
By the Schur complement formula, we have
\[
    \sfK = \sfC_{r,r}^\dag \sfC_{r,\bot} = - \sfQ_{r,\bot} \sfQ_{\bot,\bot}^\dag. 
\]
By assumption, $\sfQ = - \nabla^2\log \pi_0$ satisfies $\sfQ(i,j) = 0$ if $i \notin \mcN(j)$, so that 
\[
    \sfK(j,k) = \sum_{i \in [K]} \sfQ_{r,\bot}(j,i) \sfQ_{\bot,\bot}^\dag(i,k) = \sum_{i \in \mcN(j)} \sfQ_{r,\bot}(j,i) \sfQ_{\bot,\bot}^\dag(i,k).
\]
By \Cref{lem:MatInvLoc}, we have 
\begin{align*}
    \sum_{k\in [K]} \norm{ \sfK(j,k) } \le~& \sum_{k\in [K]} \sum_{i \in \mcN(j)} \norm{ \sfQ_{r,\bot}(j,i) } \norm{ \sfQ_{\bot,\bot}^\dag(i,k) } \\
    \le~& \sum_{i \in \mcN(j)} M \sum_{k \in [K]} \frac{1}{m} \Brac{ 1 - \frac{m}{M} }^{\sfd_\sfG(i,k)}.
\end{align*}
Since $\sfG$ is $(S,\nu)$-local, we have
\begin{align*}
    \sum_{k \in [K]} \frac{1}{m} \Brac{ 1 - \frac{m}{M} }^{\sfd_\sfG(i,k)} =~& \frac{1}{m} \sum_{q=0}^\infty \#\{ k: \sfd_\sfG(i,k) = q \} \Brac{ 1 - \frac{m}{M} }^q \\
    \le~& \frac{1}{m} \sum_{q=0}^\infty \Brac{ 1 + S q^\nu } \Brac{ 1 - \frac{m}{M} }^q \\
    =~& \frac{1}{m} \Rectbrac{ \frac{M}{m} + S \sum_{q=0}^\infty q^\nu \Brac{ 1 - \frac{m}{M} }^q } \\
    \le~& \frac{1}{m} \Rectbrac{ \frac{M}{m} + S \nu! \Brac{ \frac{m}{M} }^{-\nu-1} \Brac{ 1 - \frac{m}{M} } } \\
    =~& \frac{1}{m} \Rectbrac{ \kappa + S \nu! \kappa^{\nu+1} \Brac{ 1 - \kappa^{-1} } } \le \frac{S \nu! \kappa^{\nu+1}}{m}. 
\end{align*}
Here we use \Cref{lem:Li_ctrl} in the penultimate line. Finally, by the locality of $\sfG$, we have 
\[
    \sum_{k\in[K]} \norm{ \sfK(j,k) } \le \sum_{i \in \mcN(j)} M \cdot \frac{S \nu! \kappa^{\nu+1}}{m} \le S(1+S) \nu! \kappa^{\nu+2}. 
\]
This completes the proof. 
\end{proof}

\subsection{Proof of Theorem~6}

\begin{proof}[Proof of Theorem~6]

By \Cref{asm:LLIS} and \Cref{thm:MRF_loc}, $\pi$ is $\frac{S\nu! \kappa^\nu}{m}$-localized.
Therefore, by \Cref{thm:margin}, it holds that 
\[
    \max_{i \in [K]} \sfW_1(\pi_i,\pi_i') \le \frac{S\nu! \kappa^\nu}{m} \cdot \max_{j \in [K]} \norm{ \nabla_j \log \pi - \nabla_j \log \pi' }_{L^1(\pi')}.
\]
Recall that $\pi(\bx) \propto \ell(\bx) \pi_0(\bx)$ and $\pi'(\bx) \propto \ell_r(\bx_r) \pi_0(\bx)$, where
\[
    \log \ell_r(\bx_r) = \int \log \ell(\bx_r,\widetilde{\bx}_\bot) \pi_0(\widetilde{\bx}_\bot|\bx_r) \mdd \widetilde{\bx}_\bot.
\]
Because $\Pi_{j,\bot}\nabla_j \log \ell_r(\bx_r)=0$, we have
\begin{align*}
    \nabla_j \log \pi (\bx) \,-\, &\nabla_j \log \pi' (\bx) = \nabla_j \log \ell (\bx) - \nabla_j \log \ell_r (\bx_r) \\
    =~& \Pi_{j,\bot} \nabla_j \log \ell (\bx) + \Pi_{j,r} \nabla_j \log \ell (\bx) - \int \Pi_{j,r} \nabla_j \log \ell(\bx_r,\widetilde{\bx}_\bot) \pi_0(\widetilde{\bx}_\bot|\bx_r) \mdd \widetilde{\bx}_\bot \\
    &- \int \log \ell(\bx_r,\widetilde{\bx}_\bot) \Pi_{j,r} \nabla_j \pi_0(\widetilde{\bx}_\bot|\bx_r) \mdd \widetilde{\bx}_\bot. 
\end{align*}
Thus
\begin{equation}
\label{eqn:pf_LLIS_err_decomp}
\begin{split}
    &\norm{ \nabla_j \log \pi - \nabla_j \log \pi' }_{L^1(\pi')} \\
    \le~& \int \norm{ \Pi_{j,\bot} \nabla_j \log \ell (\bx) } \pi'(\bx) \mdd \bx \\
    &+ \int \norm{ \Pi_{j,r} \nabla_j \log \ell (\bx) - \int \Pi_{j,r} \nabla_j \log \ell(\bx_r,\widetilde{\bx}_\bot) \pi_0(\widetilde{\bx}_\bot|\bx_r) \mdd \widetilde{\bx}_\bot } \pi'(\bx) \mdd \bx \\
    &+ \int \norm{ \int \log \ell(\bx_r,\widetilde{\bx}_\bot) \Pi_{j,r} \nabla_j \pi_0(\widetilde{\bx}_\bot|\bx_r) \mdd \widetilde{\bx}_\bot } \pi'(\bx) \mdd \bx \\
    =:& \sfI_1 + \sfI_2 + \sfI_3.
\end{split}
\end{equation}
Next we control the three terms separately.

\noindent
(1) For $\sfI_1$, by the Cauchy--Schwarz inequality,
\begin{equation}
\label{eqn:pf_LLIS_I1}
\begin{split}
    \sfI_1^2 \le~& \int \norm{ \Pi_{j,\bot} \nabla_j \log \ell (\bx) }^2 \pi'(\bx) \mdd \bx \\
    =~& \int \tr \Rectbrac{  \Pi_{j,\bot} \nabla_j \log \ell (\bx) \nabla_j \log \ell(\bx) \matT \Pi_{j,\bot}\matT } \pi'(\bx) \mdd \bx \\
    =~& \tr \Rectbrac{ \Pi_{j,\bot} \Brac{  \int \nabla_j \log \ell(\bx) \nabla_j \log \ell(\bx) \matT \pi'(\bx) \mdd \bx } \Pi_{j,\bot}\matT } \\
    =~& \tr \Rectbrac{ \Pi_{j,\bot} \sfR_j \Pi_{j,\bot}\matT }.
\end{split}
\end{equation}
Recall that $\sfR_j := \int \nabla_j \log \ell(\bx) \nabla_j \log \ell(\bx) \matT \pi'(\bx) \mdd \bx$.

\ \\[-5pt]
\noindent
(2) For $\sfI_2$, similarly we have
{\small \begin{align*}
    \sfI_2^2 \le~& \int \norm{ \Pi_{j,r} \nabla_j \log \ell (\bx) - \int \Pi_{j,r} \nabla_j \log \ell(\bx_r,\widetilde{\bx}_\bot) \pi_0(\widetilde{\bx}_\bot|\bx_r) \mdd \widetilde{\bx}_\bot }^2 \pi'(\bx) \mdd \bx  \\
    =~& \int \Brac{ \int  \norm{ \Pi_{j,r} \nabla_j \log \ell (\bx) - \int \Pi_{j,r} \nabla_j \log \ell(\bx_r,\widetilde{\bx}_\bot) \pi_0(\widetilde{\bx}_\bot|\bx_r) \mdd \widetilde{\bx}_\bot }^2 \pi_0(\bx_\bot|\bx_r) \mdd \bx_\bot } \pi'(\bx_r) \mdd \bx_r.  
\end{align*}}
As a conditional distribution of $\pi_0$, $\pi_0(\cdot|\bx_r)$ is $m$-log-concave. By the Poincar\'e inequality,
\begin{align*}
    \int \norm{ \Pi_{j,r} \nabla_j \log \ell (\bx) - \int \Pi_{j,r} \nabla_j \log \ell(\bx_r,\widetilde{\bx}_\bot) \pi_0(\widetilde{\bx}_\bot|\bx_r) \mdd \widetilde{\bx}_\bot }^2 \pi_0(\bx_\bot|\bx_r) \mdd \bx_\bot & \\
    \le \frac{1}{m} \int \Fro{ \nabla_{\bx_\bot} \Brac{ \Pi_{j,r} \nabla_j \log \ell (\bx)}\matT  }^2 \pi_0(\bx_\bot|\bx_r) \mdd \bx_\bot& \\
    = \frac{1}{m} \int \Fro{ \Pi_\bot \nabla \nabla_j\matT \log \ell (\bx) \; \Pi_{j,r}\matT }^2 \pi_0(\bx_\bot|\bx_r) \mdd \bx_\bot&.
\end{align*}
Substituting this into the above inequality, we have 
\begin{equation}
\label{eqn:pf_LLIS_I2}
\begin{split}
    \sfI_2^2 \le~& \frac{1}{m} \int \Brac{ \int \Fro{ \Pi_\bot \nabla \nabla_j\matT \log \ell (\bx) \; \Pi_{j,r}\matT }^2 \pi_0(\bx_\bot|\bx_r) \mdd \bx_\bot } \pi'(\bx_r) \mdd \bx_r \\
    =~& \frac{1}{m} \int \Fro{ \Pi_\bot \nabla \nabla_j\matT \log \ell (\bx) \; \Pi_{j,r}\matT }^2 \pi'(\bx) \mdd \bx \\
    =~& \frac{1}{m} \int \tr \Rectbrac{ \Pi_\bot \nabla \nabla_j\matT \log \ell(\bx) \, \Pi_{j,r}\matT \Pi_{j,r} \, \nabla_j \nabla\matT \log \ell(\bx) \Pi_\bot\matT } \pi'(\bx) \mdd \bx \\
    \le~& \frac{\kappa}{m} \tr \Rectbrac{ \Pi_\bot \Brac{ \int \nabla \nabla_j\matT \log \ell(\bx) \, \nabla_j \nabla\matT \log \ell(\bx)\, \pi'(\bx) \mdd \bx } \Pi_\bot\matT } \\
    =~&\frac{\kappa}{m} \tr \Rectbrac{ \Pi_\bot \widetilde{\sfR}_j \Pi_\bot\matT },
\end{split}
\end{equation}
where we denote $\widetilde{\sfR}_j := \int \nabla \nabla_j\matT \log \ell(\bx)\, \nabla_j \nabla\matT \log \ell(\bx)\, \pi'(\bx) \mdd \bx$. 
Here the second inequality follows from $\Pi_{j,r}\matT \Pi_{j,r}\preceq \kappa I_{d_j}$, since $\Pi_{j,r} = \sfC_{j}^{\frac{1}{2}} \widetilde{\Pi}_{j,r} \sfC_{j}^{-\frac{1}{2}} $ for some Euclidean orthogonal projection $\widetilde{\Pi}_{j,r} := \sfC_j^{-\frac{1}{2}} \sfU_j^{} \sfU_j^{\matT} \sfC_j^{-\frac{1}{2}} $, and thus
\begin{align*}
    \forall v \in \mR^{d_j},\quad v\matT \Pi_{j,r}\matT \Pi_{j,r} v =~& v\matT \sfC_{j}^{\frac{1}{2}} \widetilde{\Pi}_{j,r} \sfC_j^{-1} \widetilde{\Pi}_{j,r} \sfC_{j}^{-\frac{1}{2}} v \le \norm{\sfC_j^{-1}} \norm{ \widetilde{\Pi}_{j,r} \sfC_{j}^{\frac{1}{2}} v }^2 \\
    \le~& \norm{ \sfC_j^{-1}} \norm{ \sfC_{j}^{\frac{1}{2}} v }^2 \le \norm{C_j^{-1}} \norm{ \sfC_{j} } \norm{ v }^2 \le \kappa \norm{v}^2.
\end{align*}
Here we use the assumption that $m I \preceq \sfC \preceq M I$, and as a principle submatrix of $\sfC$, we have $m I \preceq \sfC_j \preceq M I$ and thus $\norm{C_j^{-1}} \norm{ C_{j} } \le \kappa$.

\ \\[-5pt]
\noindent
(3) For the last term, note that
\[
    \pi_0(\widetilde{\bx}_\bot|\bx_r) = \frac{1}{Z_0} \exp \Brac{ - \frac{1}{2} \normo{ \widetilde{\bx}_\bot - \sfC_{\bot,r} \sfC_{r,r}^\dag \bx_r }^2_{\sfC_{\bot|r}}  }.
\]
Here we denote 
\[
    \sfC_{\bot,r} = \Pi_{\bot} \sfC \Pi_{r}, \quad \sfC_{r,r} = \Pi_{r} \sfC \Pi_{r}, \quad \sfC_{\bot|r} = \sfC_{\bot,\bot} - \sfC_{\bot,r} \sfC_{r,r}^\dag \sfC_{r,\bot}.
\]
Since $\pi_0(\widetilde{\bx}_\bot|\bx_r)$ depends on $\widetilde{\bx}_\bot,\bx_r$ through the term $\widetilde{\bx}_\bot - \sfC_{\bot,r} \sfC_{r,r}^\dag \bx_r$, we have
\[
    \nabla_{\bx_r} \pi_0(\widetilde{\bx}_\bot|\bx_r) = - \sfC_{r,r}^\dag \sfC_{r,\bot} \nabla_{\bx_\bot} \pi_0(\widetilde{\bx}_\bot|\bx_r). 
\]
Denote $\sfK = \sfC_{r,r}^\dag \sfC_{r,\bot}$. By integration by parts, we have 
\begin{align*}
    \int \log \ell(\bx_r,\widetilde{\bx}_\bot) \nabla_{\bx_r} \pi_0(\widetilde{\bx}_\bot|\bx_r) \mdd \widetilde{\bx}_\bot =~& - \sfK \int \log \ell(\bx_r,\widetilde{\bx}_\bot) \nabla_{\bx_\bot} \pi_0(\widetilde{\bx}_\bot|\bx_r) \mdd \widetilde{\bx}_\bot \\
    =~& \sfK \int \nabla_{\bx_\bot} \log \ell(\bx_r,\widetilde{\bx}_\bot) \pi_0(\widetilde{\bx}_\bot|\bx_r) \mdd \widetilde{\bx}_\bot.
\end{align*}
The $j$-th block of the above inequality reads (note $\nabla_{\bx_{j,r}} f = \Pi_{j,r} \nabla_j f$ and $\nabla_{\bx_{k,\bot}} f = \Pi_{k,\bot} \nabla_k f$)
\begin{align*}
    & \norm{ \int \log \ell(\bx_r,\widetilde{\bx}_\bot) \Pi_{j,r} \nabla_j \pi_0(\widetilde{\bx}_\bot|\bx_r) \mdd \widetilde{\bx}_\bot } \\
    \le~& \sum_{k\in[K]} \norm{\sfK(j,k)} \norm{ \int \Pi_{k,\bot} \nabla_k \log \ell(\bx_r,\widetilde{\bx}_\bot) \pi_0(\widetilde{\bx}_\bot|\bx_r) \mdd \widetilde{\bx}_\bot } \\
    \le~& \sum_{k\in[K]} \norm{\sfK(j,k)} \int \norm{ \Pi_{k,\bot} \nabla_k \log \ell(\bx_r,\widetilde{\bx}_\bot) }\pi_0(\widetilde{\bx}_\bot|\bx_r) \mdd \widetilde{\bx}_\bot. 
\end{align*}
Since $\pi'(\bx) = \pi'(\bx_\bot|\bx_r) \pi'(\bx_r) = \pi_0(\bx_\bot|\bx_r) \pi'(\bx_r)$, we have
\begin{align*}
    \sfI_3 =~& \int \norm{ \int \log \ell(\bx_r,\widetilde{\bx}_\bot) \Pi_{j,r} \nabla_j \pi_0(\widetilde{\bx}_\bot|\bx_r) \mdd \widetilde{\bx}_\bot } \pi'(\bx) \mdd \bx \\
    \le~& \int \sum_{k\in[K]} \norm{\sfK(j,k)} \Brac{ \int \norm{ \Pi_{k,\bot} \nabla_k \log \ell(\bx_r,\widetilde{\bx}_\bot) } \pi_0(\widetilde{\bx}_\bot|\bx_r) \mdd \widetilde{\bx}_\bot } \pi'(\bx_r) \mdd \bx_r \\
    =~& \sum_{k\in[K]} \norm{\sfK(j,k)} \int \norm{ \Pi_{k,\bot} \nabla_k \log \ell(\bx) } \pi'(\bx) \mdd \bx. 
\end{align*}
As in the control of $\sfI_1$, this term can be controlled by
\[
    \sfI_3 \le \sum_{k\in[K]} \norm{\sfK(j,k)} \Brac{ \tr \Rectbrac{ \Pi_{k,\bot} \sfR_k \Pi_{k,\bot}\matT } }^{1/2}.
\]
By \Cref{lem:Kl1Bound}, $\sum_{k\in[K]} \norm{\sfK(j,k)} \le S(1+S) \nu! \kappa^{\nu+2}$. Therefore,
\begin{equation}
\label{eqn:pf_LLIS_I3}
    \sfI_3 \le S(1+S) \nu! \kappa^{\nu+2}\, \max_{k\in[K]} \Brac{ \tr \Rectbrac{ \Pi_{k,\bot} \sfR_k \Pi_{k,\bot}\matT } }^{1/2}.
\end{equation}
Substituting \eqref{eqn:pf_LLIS_I1}, \eqref{eqn:pf_LLIS_I2}, and \eqref{eqn:pf_LLIS_I3} into \eqref{eqn:pf_LLIS_err_decomp}, we have
\begin{align*}
    & \max_{j\in[K]} \norm{ \nabla_j \log \pi - \nabla_j \log \pi' }_{L^1(\pi')} \\
    \le~& \Brac{ 1 + S(1+S) \nu! \kappa^{\nu+2} } \max_{k\in[K]} \Brac{ \tr \Rectbrac{ \Pi_{k,\bot} \sfR_k \Pi_{k,\bot}\matT } }^{1/2} + \Brac{\frac{\kappa}{m}}^{1/2} \max_{k\in[K]} \Brac{ \tr \Rectbrac{ \Pi_\bot \widetilde{\sfR}_k \Pi_\bot\matT } }^{1/2}.
\end{align*}
Therefore, by the marginal transport inequality, we have 
\[
    \max_{i\in[K]} \sfW_1(\pi_i,\pi_i') \le C \Brac{ \max_{j\in[K]} \tr \Rectbrac{ \Pi_{j,\bot} \sfR_j \Pi_{j,\bot}\matT } + \max_{j\in[K]}  \tr \Rectbrac{ \Pi_\bot \widetilde{\sfR}_j \Pi_\bot\matT } }^{1/2},
\]
for some dimension-independent constant $C$ that depends on $S,\nu,\kappa,m$ only.
\end{proof}

\section{Proof of Theorem~7}
\label{app:LSM}

\begin{proof}
By \eqref{eqn:hypo_class}--\eqref{eqn:U_j}, every $\pi_\theta\in\msP$ has a dependency graph compatible with $\sfG$ and satisfies the uniform Hessian bounds in \Cref{thm:MRF_loc}. Hence every $\pi_\theta\in\msP$ is $\delta$-localized with $\delta = \frac{S \nu! \kappa^\nu }{m}$; in particular, this holds for $\widehat\pi=\pi_{\widehat{\theta}_*}$.
\Cref{thm:margin} and the Cauchy--Schwarz inequality therefore give
\[
    \max_{i \in \sfV} \sfW_1(\pi_i,\widehat{\pi}_i)\le \delta\, \max_{j\in \sfV} \norm{ \nabla_j \log \pi - \nabla_j \log \widehat{\pi} }_{L^1(\pi)} \le \delta\, \max_{j\in \sfV} \normo{ \nabla_j \log \pi - \nabla_j \log \widehat{\pi} }_{L^2(\pi)}.
\]
Since $\pi \in \msP$, there exists $\theta^* \in \Theta$ such that $\pi_{\theta^*} = \pi$. Therefore, by the local score-matching identity following \eqref{eqn:J_j},
\begin{equation}
\label{eqn:pf_LSM_Jloc}
\begin{split}
    J_{\rm loc}(\theta) :=~& \max_{j \in \sfV}\, ( J_j(\theta) - \inf_{\phi \in \Theta} J_j(\phi) )
    = \max_{j \in \sfV}\, ( J_j(\theta) - J_j(\theta^*) ) \\
    =~& \max_{j \in \sfV}\, \E_{\bx \sim \pi} \normo{ \nabla_j \log \pi - \nabla_j \log \pi_{\theta} }^2.
\end{split}
\end{equation}
Substituting \eqref{eqn:pf_LSM_Jloc} into the marginal transport inequality and recalling that $\widehat{\pi} := \pi_{\widehat{\theta}_*}$, we obtain
\begin{equation}
\label{eqn:pf_LSM_step1}
    \max_{i \in \sfV} \sfW_1(\pi_i,\widehat{\pi}_i) \le \delta\, J_{\rm loc}(\widehat{\theta}_*)^{1/2}.
\end{equation}
To match the compact-domain truncation in the empirical loss, define the truncated population loss
\begin{equation}
\label{eqn:pf_LSM_Jloc_hat}
    J_j^\Lambda(\theta) := \E_{\bx \sim \pi} \Rectbrac{ \Brac{ 2 \tr \Brac{ \nabla_j s_{\theta,j} (\bx) } + \norm{s_{\theta,j}(\bx)}^2 } \ind_{\Omega_j^\Lambda} (\bx) }.
\end{equation}
Recall that $\Omega_j^\Lambda := \left\{ \bx\in\mR^d : \bx_{\mcN(j)}\in[-\Lambda,\Lambda]^{d_{\mcN(j)}} \right\}$.
Likewise, define
\begin{equation}
\label{eqn:pf_LSM_Jloc_hat_emp}
    J_{\rm loc}^\Lambda(\theta) := \max_{j \in \sfV}\, ( J_j^\Lambda(\theta) - \inf_{\theta \in \Theta} J_j^\Lambda(\theta) ).
\end{equation}
Using $J_{\rm loc}(\theta^*) = 0$ and the optimization guarantee \eqref{eqn:theta_emp}, the standard error decomposition gives
\begin{equation}
\label{eqn:pf_LSM_err_decomp}
\begin{split}
    J_{\rm loc} (\widehat{\theta}_*) =~& \Brac{ J_{\rm loc} (\widehat{\theta}_*) - J_{\rm loc}^\Lambda (\widehat{\theta}_*) } + \Brac{ J_{\rm loc}^\Lambda (\widehat{\theta}_*) - \widehat{J}_{\rm loc}^\Lambda (\widehat{\theta}_*) } + \Brac{ \widehat{J}_{\rm loc}^\Lambda(\widehat{\theta}_*) - \widehat{J}_{\rm loc}^\Lambda(\theta^*) } \\
    &+ \Brac{ \widehat{J}_{\rm loc}^\Lambda (\theta^*) - J_{\rm loc}^\Lambda (\theta^*) } + \Brac{ J_{\rm loc}^\Lambda (\theta^*) - J_{\rm loc} (\theta^*) } \\
    \le~& 2 \sup_{\theta \in \Theta} \normeo{ J_{\rm loc}(\theta) - J_{\rm loc}^\Lambda(\theta) } + 2 \sup_{\theta \in \Theta} \normeo{ \widehat{J}_{\rm loc}^\Lambda(\theta) - J_{\rm loc}^\Lambda(\theta) } + \varepsilon_{\rm opt}.
\end{split}
\end{equation}
By the definitions of $J_j^\Lambda$ in \eqref{eqn:pf_LSM_Jloc_hat} and $J_{\rm loc}$ in \eqref{eqn:J_minmax},
\begin{align*}
    \sup_{\theta \in \Theta} \normeo{ J_{\rm loc}(\theta) - J_{\rm loc}^\Lambda(\theta) } =~& \sup_{\theta \in \Theta} \normeo{ \max_{j \in \sfV}\, ( J_j(\theta) - \inf_{\theta \in \Theta} J_j(\theta) ) - \max_{j \in \sfV}\, ( J_j^\Lambda(\theta) - \inf_{\theta \in \Theta} J_j^\Lambda(\theta) ) } \\
    \le~& \sup_{\theta \in \Theta} \max_{j \in \sfV} \Brac{ \normeo{ J_j(\theta) - J_j^\Lambda(\theta) } + \normeo{ \inf_{\theta \in \Theta} J_j(\theta)- \inf_{\theta \in \Theta} J_j^\Lambda(\theta) } } \\
    \le~& 2 \sup_{\theta \in \Theta} \max_{j \in \sfV} \normeo{ J_j(\theta) - J_j^\Lambda(\theta) } =: 2 \varepsilon_{\rm tr}^\Lambda.
\end{align*}
Similarly, by the definitions of $J_{\rm loc}^\Lambda$ in \eqref{eqn:pf_LSM_Jloc_hat_emp} and $\widehat{J}_{\rm loc}^\Lambda$ in \eqref{eqn:J_loc_emp},
\begin{equation}
\label{eqn:pf_LSM_J_diff}
\begin{split}
    \sup_{\theta \in \Theta} \normeo{ J_{\rm loc}^\Lambda (\theta) - \widehat{J}_{\rm loc}^\Lambda(\theta) } =~& \sup_{\theta \in \Theta} | \max_{j \in \sfV}\, ( J_j^\Lambda(\theta) - \inf_{\theta \in \Theta} J_j^\Lambda(\theta) ) - \max_{j \in \sfV}\, ( \widehat{J}_j^\Lambda(\theta) - \inf_{\theta \in \Theta} \widehat{J}_j^\Lambda(\theta) ) | \\
    \le~& \sup_{\theta \in \Theta} \max_{j \in \sfV} \Brac{ \normeo{ \widehat{J}_j^\Lambda(\theta) - J_j^\Lambda(\theta) } + \normeo{ \inf_{\theta \in \Theta} J_j^\Lambda(\theta)- \inf_{\theta \in \Theta} \widehat{J}_j^\Lambda(\theta) } } \\
    \le~& 2 \sup_{\theta \in \Theta} \max_{j \in \sfV} \normeo{ \widehat{J}_j^\Lambda(\theta) - J_j^\Lambda(\theta) }.
\end{split}
\end{equation}
Combining \eqref{eqn:pf_LSM_step1}, \eqref{eqn:pf_LSM_err_decomp}, and \eqref{eqn:pf_LSM_J_diff} gives
\begin{equation}
\label{eqn:pf_LSM_step2}
    \max_{i \in \sfV} \sfW_1(\pi_i,\widehat{\pi}_i) \le \delta \Brac{ 4 \sup_{\theta \in \Theta} \max_{j \in \sfV} \normeo{ \widehat{J}_j^\Lambda(\theta) - J_j^\Lambda(\theta) } + 4 \varepsilon_{\rm tr}^\Lambda + \varepsilon_{\rm opt} }^{1/2}.
\end{equation}
We first control the truncation error $\varepsilon_{\rm tr}^\Lambda$.
The sparsity of the Hessian implies that $s_{\theta,j}(\bx)=\nabla_j\log\pi_\theta(\bx)$ depends only on $\bx_{\mcN(j)}$.
At the origin,
\[
    \norm{s_{\theta,j}(0)} = \norm{ \sum_{k\in\mcN(j)} \nabla_j u_{\theta,k}(0) } \le \sum_{k\in\mcN(j)} \norm{\nabla_j u_{\theta,k}(0)} \le (S+1)R,
\]
where we used $|\mcN(j)|\le S+1$ and the $C^\gamma$ bound in \eqref{eqn:U_j}.
Since $\norm{\nabla s_{\theta,j}(\bx)}=\norm{\nabla\nabla_j\matT\log\pi_\theta(\bx)}\le M$, it follows that, for any $\bx$,
\[
    \norm{ s_{\theta,j}(\bx) } \le \norm{ s_{\theta,j}(0) } + M \norm{\bx_{\mcN(j)}} \le (S+1)R + M \norm{\bx_{\mcN(j)}},
\]
\[
    \norme{ \tr \Brac{ \nabla_j s_{\theta,j}(\bx) } } = \norme{ \tr \Brac{ \nabla_{jj}^2 \log \pi_\theta(\bx) } } \le M d_j.
\]
Consequently, uniformly over $\theta\in\Theta$ and $j\in\sfV$,
\[
    \normeo{2\tr\Brac{\nabla_j s_{\theta,j}(\bx)}+\norm{s_{\theta,j}(\bx)}^2}
    \le C_0\Brac{1+\norm{\bx_{\mcN(j)}}^2},
\]
where $C_0$ depends only on $S,M,R$, and $d_{\rm loc}$.
Since $\pi$ is $m$-strongly log-concave, its marginal distribution on $\bx_{\mcN(j)}$ is also $m$-strongly log-concave and hence has Gaussian tails. Together with the uniform local score and Hessian bounds above, this implies
\[
    \mP_{\bx\sim\pi}\Brac{\normo{\bx_{\mcN(j)}}>t}\le C_1\mee^{-mt^2/4}, \qquad t>0,
\]
where $C_1$ depends only on $S,\nu,m,M,R$, and $d_{\rm loc}$, and is therefore uniform in $j$ and independent of the ambient dimension $d$.
Since $(\Omega_j^\Lambda)\matc\subseteq\{\norm{\bx_{\mcN(j)}}>\Lambda\}$, integration of this tail bound gives
\begin{align*}
    \normeo{J_j(\theta)-J_j^\Lambda(\theta)}
    &\le C_0 \E_{\bx\sim\pi} \Rectbrac{\Brac{1+\normo{\bx_{\mcN(j)}}^2}\ind_{\{\normo{\bx_{\mcN(j)}}>\Lambda\}}} \\
    &= C_0(1+\Lambda^2) \mP_{\bx\sim\pi} \Brac{ \normo{\bx_{\mcN(j)}}>\Lambda } 
    +2C_0\int_\Lambda^\infty t\mP_{\bx\sim\pi} \Brac{\normo{\bx_{\mcN(j)}}>t} \mdd t \\
    &\le C_0C_1\Brac{1+\Lambda^2+\frac{4}{m}}\mee^{-m\Lambda^2/4}
    \le C'(1+\Lambda^2)\mee^{-m\Lambda^2/4}.
\end{align*}
Therefore, by definition,
\begin{equation}
\label{eqn:pf_LSM_truncation_error}
    \varepsilon_{\rm tr}^\Lambda = \sup_{\theta \in \Theta} \max_{j \in \sfV} \normeo{ J_j(\theta) - J_j^\Lambda(\theta) } \le C'(1+\Lambda^2) \mee^{-m\Lambda^2/4} \le C'' \mee^{-m\Lambda^2/8}.
\end{equation}
Here we used $1+\Lambda^2 \le C \mee^{m\Lambda^2/8}$ for a constant $C$ depending only on $m$.

It remains to control $\sup_{\theta \in \Theta} \max_{j \in \sfV} \normeo{ \widehat{J}_j^\Lambda(\theta) - J_j^\Lambda(\theta) }$.
Given data $\bX^{(1)},\dots,\bX^{(N)}$, denote
\[
    \mP_N f := \frac{1}{N} \sum_{i=1}^N f(\bX^{(i)}),\quad f_{\theta,j}^\Lambda(\bx) := \Brac{2 \tr \Brac{ \nabla_j s_{\theta,j} (\bx) } + \norm{s_{\theta,j}(\bx)}^2}\ind_{\Omega_j^\Lambda}(\bx).
\]
Then, by definition,
\begin{equation}
\label{eqn:pf_LSM_emp}
    \widehat{J}_j^\Lambda(\theta) - J_j^\Lambda(\theta) = (\mP_N - \E_\pi) f_{\theta,j}^\Lambda.
\end{equation}
This term can be controlled by standard empirical process techniques \cite{MR1385671}.
We provide the details below for completeness.

\ \\[-5pt]
\noindent
(I) Lipschitz control of $f_{\theta,j}^\Lambda$ by the local potentials.
Denote $\Delta_j u := \tr [\nabla_{jj}^2 u] $.
Because the graph-sparsity condition on the local potentials in \eqref{eqn:U_j} makes $s_{\theta,j}$ a function of $\bx_{\mcN(j)}$ only, on $\Omega_j^\Lambda$ we may evaluate the sum defining $s_{\theta,j}$ at the representative point obtained by setting all coordinates outside $\mcN(j)$ to zero.
All arguments of the relevant potential functions then lie in their respective truncation cubes, so the $C^\gamma$ bound in \eqref{eqn:U_j} applies.
Therefore, for any $\theta,\phi \in \Theta$,
\begin{align*}
    | f_{\theta,j}^\Lambda - f_{\phi,j}^\Lambda| \le~& \Brac{2 | \Delta_j \log \pi_\theta - \Delta_j \log \pi_\phi | + | \norm{ \nabla_j \log \pi_\theta }^2 - \norm{ \nabla_j \log \pi_\phi }^2 |}\ind_{\Omega_j^\Lambda} \\
    \le~& 2 | \Delta_j \sum_{k\in \mcN(j)} \Brac{ u_{\theta,k} - u_{\phi,k} } | + \norm{ \nabla_j \sum_{k\in \mcN(j)} \Brac{ u_{\theta,k} + u_{\phi,k} } } \norm{ \nabla_j \sum_{k\in \mcN(j)} \Brac{ u_{\theta,k} - u_{\phi,k} } } \\
    \le~& 2 \sum_{k\in \mcN(j)} | \Delta_j \Brac{ u_{\theta,k} - u_{\phi,k} } | + \sum_{k\in \mcN(j)} \norm{ \nabla_j \Brac{ u_{\theta,k} + u_{\phi,k} } } \cdot \sum_{k\in \mcN(j)} \norm{ \nabla_j \Brac{ u_{\theta,k} - u_{\phi,k} } } \\
    \le~& 2 \sum_{k\in \mcN(j)} \Big( | \Delta_j \Brac{ u_{\theta,k} - u_{\phi,k} } | + (S + 1) R \cdot \norm{ \nabla_j \Brac{ u_{\theta,k} - u_{\phi,k} } }  \Big) \\
    \le~& C \sum_{k\in \mcN(j)} \norm{ u_{\theta,k} - u_{\phi,k} }_{C^2}.
\end{align*}
Here and below, $C$ denotes a positive constant depending only on $S,R$, and $d_{\rm loc}$; its value may change from line to line. In the fourth line, we used $|\mcN(j)| \le S + 1$ and $\norm{ \nabla_j u_{\theta,k} } \le R$.

\ \\[-5pt]
\noindent
(II) $\varepsilon$-net argument.
For fixed $j \in \sfV$, consider the vector function space 
\[
    \mcV_j := \left\{ \bu_\theta = \left(u_{\theta,k}|_{[-\Lambda,\Lambda]^{d_{\mcN(k)}}}\right)_{k \in \mcN(j)}: u_{\theta,k} \in \mcU_k \right\}.
\]
By \eqref{eqn:U_j}, each restriction $u_{\theta,k}|_{[-\Lambda,\Lambda]^{d_{\mcN(k)}}}$ belongs to the radius-$R$ ball in $C^\gamma([-\Lambda,\Lambda]^{d_{\mcN(k)}})$.
Equip $\mcV_j$ with norm 
\[
    \norm{ \bu_\theta }_{\mcV_j} := \max_{k\in \mcN(j)} \norm{ u_{\theta,k} }_{C^2([-\Lambda,\Lambda]^{d_{\mcN(k)}})}.
\]
By definition, $\sup_{\bu_\theta \in \mcV_j} \norm{\bu_\theta}_{\mcV_j} \le R$.
Consider an $\varepsilon$-net $\mcV_j^\varepsilon = \{ \bu_{\theta_\alpha} \}_{\alpha \in [N_j^\varepsilon]} \subset \mcV_j$ under $\norm{\cdot}_{\mcV_j}$, i.e.,
\[
    \forall \bu_\theta \in \mcV_j, \quad \exists \bu_{\theta_\alpha} \in \mcV_j^\varepsilon, \quad \st\norm{ \bu_\theta - \bu_{\theta_\alpha} }_{\mcV_j} \le \varepsilon.
\]
By the Lipschitz control in (I), for any $\theta \in \Theta$, there exists some $\alpha \in [N_j^\varepsilon]$ such that
\[
    \norm{ f_{\theta,j}^\Lambda - f_{\theta_\alpha,j}^\Lambda }_{L^\infty} \le C \sum_{k\in \mcN(j)} \norm{ u_{\theta,k} - u_{\theta_\alpha,k} }_{C^2} \le C\varepsilon.
\]
Here we use $|\mcN(j)| \le S + 1$ again.
Therefore, by \eqref{eqn:pf_LSM_emp}, for every $\theta \in \Theta$ there exists $\alpha \in [N_j^\varepsilon]$ such that
\begin{align*}
    | \widehat{J}_j^\Lambda(\theta) - J_j^\Lambda(\theta) | =~& | (\mP_N - \E_\pi) f_{\theta,j}^\Lambda | \le | (\mP_N - \E_\pi) f_{\theta_\alpha,j}^\Lambda | + | (\mP_N - \E_\pi)\Brac{ f_{\theta,j}^\Lambda - f_{\theta_\alpha,j}^\Lambda } | \\
    \le~& | (\mP_N - \E_\pi) f_{\theta_\alpha,j}^\Lambda | + C\varepsilon.
\end{align*}
This further implies that for any $\varepsilon >0$,
\begin{equation}
\label{eqn:pf_LSM_eps_bound}
    \sup_{\theta \in \Theta} | \widehat{J}_j^\Lambda(\theta) - J_j^\Lambda(\theta) | \le \max_{1\le \alpha \le N_j^\varepsilon} | (\mP_N - \E_\pi) f_{\theta_\alpha,j}^\Lambda | + C\varepsilon.
\end{equation}

\ \\[-5pt]
\noindent
(III) Covering number bound.
Let $\msN(\varepsilon,\mcF,\norm{\cdot})$ denote the covering number of a function class $\mcF$ under $\norm{\cdot}$, i.e., the minimal number of $\varepsilon$-balls needed to cover $\mcF$.
By definition, we have $N_j^\varepsilon \le \msN(\varepsilon,\mcV_j,\norm{\cdot}_{\mcV_j})$.
For each $k\in\mcN(j)$, the classical metric entropy estimate for smooth functions \cite{MR1385671}, applied on a cube of side length $2\Lambda$, gives
\begin{equation}
\label{eqn:Pf_entropy_L}
\begin{split}
    &\log \msN\Brac{\varepsilon,
    \left\{u\in C^\gamma([-\Lambda,\Lambda]^{d_{\mcN(k)}}):\norm{u}_{C^\gamma}\le R\right\},
    \norm{\cdot}_{C^2}} \\
    &\qquad\le C_{d_{\mcN(k)},\gamma}\Lambda^{d_{\mcN(k)}}(R/\varepsilon)^{d_{\mcN(k)}/(\gamma-2)},
    \qquad 0<\varepsilon\le R.
\end{split}
\end{equation}
The constant $C_{d_{\mcN(k)},\gamma}$ is independent of $\Lambda$, $R$, and $\varepsilon$.
By construction, $\mcV_j$ is contained in the product of these radius-$R$ balls, equipped with the $\ell_\infty$-norm $\norm{u}_{\mcV_j} = \max_{k\in \mcN(j)} \norm{u_k}_{C^2}$.
Therefore, by \eqref{eqn:Pf_entropy_L} and standard properties of covering numbers of product spaces, we have
\begin{equation}
\label{eqn:Pf_NBound}
    \log N_j^\varepsilon
    \le \sum_{k \in \mcN(j)} C_{d_{\mcN(k)},\gamma}\Lambda^{d_{\mcN(k)}}(R/\varepsilon)^{d_{\mcN(k)}/(\gamma-2)}
    \le (S+1)C_{d_{\rm loc},\gamma}\Lambda^{d_{\rm loc}}(R/\varepsilon)^{d_{\rm loc}/(\gamma-2)}.
\end{equation}
Recall that $d_{\rm loc} = \max_{j\in\sfV} d_{\mcN(j)}$.
This bound is uniform for all $j \in \sfV$.

\ \\[-5pt]
\noindent
(IV) Concentration inequality.
First note that $f_{\theta,j}^\Lambda$ is uniformly bounded:
\[
    \norm{f_{\theta,j}^\Lambda}_{L^\infty} \le 2d_j(S+1)R+(S+1)^2R^2 \le C_{\rm env}.
\]
By Hoeffding's inequality, for any fixed $\alpha \in [N_j^\varepsilon]$,
\[
    \mP \Brac{ | (\mP_N - \E_\pi) f_{\theta_\alpha,j}^\Lambda | \ge t } \le 2\mee^{-Nt^2/(2C_{\rm env}^2)}, \quad \forall t > 0.
\]
Therefore, by the union bound,
\begin{align*}
    \mP \Brac{ \max_{1\le \alpha\le N_j^\varepsilon} | (\mP_N - \E_\pi) f_{\theta_\alpha,j}^\Lambda | \ge t } \le~& 2N_j^\varepsilon \mee^{-Nt^2/(2C_{\rm env}^2)} \\
    \le~& 2\exp \Brac{  (S+1) C_{d_{\rm loc},\gamma}\Lambda^{d_{\rm loc}} (R/\varepsilon)^{d_{\rm loc}/(\gamma-2)} - \frac{N t^2}{2C_{\rm env}^2} }.
\end{align*}
Applying the union bound again over $j \in \sfV$ gives
\begin{equation}
\label{eqn:pf_LSM_max_ineq}
\begin{split}
    \mP \Brac{ \max_{j\in \sfV} \max_{1\le \alpha\le N_j^\varepsilon} | (\mP_N - \E_\pi) f_{\theta_\alpha,j}^\Lambda | \ge t } \le |\sfV| \cdot \max_j \mP \Brac{ \max_{1\le \alpha\le N_j^\varepsilon} | (\mP_N - \E_\pi) f_{\theta_\alpha,j}^\Lambda | \ge t } &\\
    \le 2\exp \Brac{ \log |\sfV| + (S+1) C_{d_{\rm loc},\gamma}\Lambda^{d_{\rm loc}} (R/\varepsilon)^{d_{\rm loc}/(\gamma-2)} - \frac{N t^2}{2C_{\rm env}^2} } &.
\end{split}
\end{equation}

\ \\[-5pt]
\noindent
(V) Conclusion.
Taking $t=C\varepsilon$ in \eqref{eqn:pf_LSM_max_ineq}, with $C$ large enough to absorb the final term in \eqref{eqn:pf_LSM_eps_bound}, and using \eqref{eqn:pf_LSM_step2} and \eqref{eqn:pf_LSM_truncation_error}, we have for any $0<\varepsilon\le R$,
\begin{align*}
    &\mP \Brac{ \max_{i \in \sfV} \sfW_1(\pi_i,\widehat{\pi}_i) \ge \delta\Brac{8t+4C''\mee^{-m\Lambda^2/8}+\varepsilon_{\rm opt}}^{1/2} } \\
    \le~& \mP \Brac{ \max_{j\in \sfV} \max_{1\le \alpha\le N_j^\varepsilon} | (\mP_N - \E_\pi) f_{\theta_\alpha,j}^\Lambda | \ge t } \\
    \le~& 2\exp \Brac{ \log |\sfV| + (S+1) C_{d_{\rm loc},\gamma}\Lambda^{d_{\rm loc}} (R/\varepsilon)^{d_{\rm loc}/(\gamma-2)} - \frac{N \varepsilon^2}{2} }.
\end{align*}
To make the right-hand side at most $\beta \in (0,1)$, it suffices that
\[
    N \ge C\max \Brac{ \varepsilon^{-2} \log (|\sfV|/\beta) ,
    \Lambda^{d_{\rm loc}}\varepsilon^{ - (d_{\rm loc}+2(\gamma-2))/(\gamma-2) } }.
\]
Therefore, with probability at least $1-\beta$ (noting that $|\sfV| \le d$),
\begin{align*}
    \max_{i \in \sfV} \sfW_1(\pi_i,\widehat{\pi}_i)
    \le~& C \max \left\{ \left(\frac{\log(d/\beta)}{N}\right)^{1/4}, \left(\frac{\Lambda^{d_{\rm loc}}}{N}\right)^{(\gamma-2)/(2(d_{\rm loc} + 2(\gamma-2)))} \right\} \\
    &+ C\mee^{-m\Lambda^2/16} + \delta\varepsilon_{\rm opt}^{1/2}.
\end{align*}
Here $C$ depends on $d_{\rm loc},\gamma,R,S,\nu,m$, and $M$. This concludes the proof.
\end{proof}

\bibliographystyle{siam}
\bibliography{BibLib}

\end{document}